\documentclass{article}

\PassOptionsToPackage{numbers, compress}{natbib}

\usepackage[preprint]{neurips_2026}

\usepackage[utf8]{inputenc} 
\usepackage[T1]{fontenc}    
\PassOptionsToPackage{hypertexnames=false}{hyperref}
\usepackage{hyperref}       
\usepackage{url}            
\hypersetup{
	pdftitle={Decoupling Direction and Magnitude: Language-Steered Flow Matching for Super-Resolution in the Dark},
	pdfauthor={Jiaxin Gao, Ziyu Yue, Yaohua Liu, Danchen Cui, Zhihui Zhao, and Zhixun Su}
}
\usepackage{booktabs}       
\usepackage{amsfonts}       
\usepackage{nicefrac}       
\usepackage{microtype}      
\usepackage{booktabs}       
\usepackage{amsmath}        
\usepackage{amssymb}        
\usepackage{amsfonts}       
\usepackage{nicefrac}       

\usepackage{algorithm}
\usepackage{algorithmic}

\usepackage[table]{xcolor}
\usepackage{booktabs}
\usepackage{multirow}
\usepackage{graphicx}
\usepackage{bm}
\usepackage{array}
\usepackage{placeins}

\usepackage{pifont}
\newcommand{\cmark}{\ding{51}}
\newcommand{\xmark}{\ding{55}}

\newcommand{\srtype}{\(\blacktriangle\)}
\newcommand{\difftype}{\(\blacklozenge\)}
\newcommand{\castype}{\(\clubsuit\)}

\definecolor{rankone}{RGB}{255,218,218}
\definecolor{ranktwo}{RGB}{255,229,190}
\definecolor{rankthree}{RGB}{255,248,190}

\newcommand{\promptslot}[1]{\colorbox{blue!10}{\textcolor{blue!65!black}{\textbf{#1}}}}

\definecolor{tabhead}{RGB}{242,246,250}
\definecolor{pairrow}{RGB}{250,252,255}
\definecolor{inputrow}{RGB}{246,250,246}
\definecolor{slotrow}{RGB}{255,250,242}

\newcommand{\pairedtag}{\textcolor{blue!65!black}{\textbf{P}}}
\newcommand{\inputtag}{\textcolor{green!45!black}{\textbf{I}}}
\newcommand{\prompttag}{\textcolor{orange!80!black}{\textbf{T}}}

\definecolor{losshead}{RGB}{242,246,250}
\definecolor{losslatent}{RGB}{238,246,255}
\definecolor{lossrecon}{RGB}{246,250,246}
\definecolor{losscontrol}{RGB}{255,248,238}

\definecolor{condhead}{RGB}{242,246,250}
\definecolor{conddeploy}{RGB}{236,248,238}
\definecolor{condgeneric}{RGB}{248,249,251}
\definecolor{condoracle}{RGB}{255,248,238}

\newcommand{\nogt}{\textcolor{green!50!black}{\textbf{No}}}
\newcommand{\yesgt}{\textcolor{orange!85!black}{\textbf{Yes}}}

\newcommand{\appref}[1]{\textbf{Appx.~\ref{#1}}}

\title{Decoupling Direction and Magnitude: Language-Steered Flow Matching for Super-Resolution in the Dark}
\author{%
  \normalfont Jiaxin Gao$^{1}$ \quad Ziyu Yue$^{2}$ \quad Yaohua Liu$^{3,*}$ \\
  Danchen Cui$^{4}$ \quad Zhihui Zhao$^{4}$ \quad Zhixun Su$^{2}$ \\
  $^{1}$The Hong Kong Polytechnic University \quad
  $^{2}$Dalian University of Technology \\
  $^{3}$The University of Hong Kong \quad
  $^{4}$Taiyuan University of Technology \\
  $^{*}$Corresponding author: \texttt{liuyaohua.918@gmail.com}
}

\begin{document}

\maketitle
 
 \begin{abstract}
 	 Image super-resolution \textit{in the dark} is fundamentally challenged by extreme spatial heterogeneity: severely underexposed and noisy regions demand aggressive generative enhancement to reconstruct missing details, while relatively well-exposed areas require conservative updates to prevent hallucinated textures. Standard diffusion and flow-matching models, however, rely on globally uniform integration trajectories, making them sub-optimal for such spatially varying degradations. In this paper, we propose \textbf{DeLang-SR} (\textbf{De}gradation-\textbf{Lang}uage steered \textbf{S}uper-\textbf{R}esolution), a one-step flow-matching framework that explicitly decouples the restoration \textit{direction} and \textit{magnitude}. Rather than treating low-light corruption as arbitrary latent features, DeLang-SR translates measurable physical statistics into structured degradation language, harnessing the semantic prior of pretrained text-to-image models. The degradation-language intent, complemented by image-specific visual tokens, steers the velocity field (direction) toward restoration-relevant manifolds. Simultaneously, a continuous spatial update-strength field acts as a locally varying Euler step-size field (magnitude), applying stronger integration steps in heavily degraded shadows while enforcing conservative updates in reliable regions. Experiments on paired and real dark images, together with ablations and diagnostic analyses, show that DeLang-SR provides an effective and interpretable one-step route to super-resolution in the dark. Code and pretrained models are available at \url{https://github.com/moriyaya/DeLang-SR}.
 \end{abstract}

 \section{Introduction}
 Generative restoration models~\cite{saharia2022palette, wang2023zero, fei2023generative} are often built upon an implicit simplification: a single conditioned trajectory is expected to repair an entire degraded image. This simplification becomes problematic for image super-resolution \textit{in the dark}~\cite{aakerberg2021rellisur,chen2018learning}. A dark low-resolution image is not uniformly corrupted. Some regions are dominated by underexposure and shadow noise, where visibility and structures are severely suppressed; other regions remain relatively reliable and should stay close to the observation. Applying the same restoration strength everywhere is therefore intrinsically risky: weak updates fail in shadows, while strong updates hallucinate textures or distort reliable details. Dark super-resolution thus exposes a clear mismatch between globally driven restoration trajectories and spatially heterogeneous degradations.
 
 Existing low-light enhancement and super-resolution methods usually address this challenge through cascaded pipelines or task-specific architectures~\cite{jiang2021enlightengan, zamir2022restormer,yue2024unveiling}. While effective in improving brightness and resolution, they are often constrained by deterministic reconstruction objectives and limited generative capability under severe degradation. Pretrained diffusion and flow-matching models~\cite{wu2025omgsr,Qin2025ReversingFF} offer a powerful alternative by injecting rich natural-image priors into restoration. However, these priors are usually activated through generic prompts, degradation-agnostic conditions, or globally shared sampling schedules~\cite{wang2023promptir, luo2024daclip}. Such mechanisms can indicate that an image should be restored, but they do not explicitly answer two more important questions: \emph{toward which restoration direction should the latent state move, and how far should each region move?}
 
 We revisit dark super-resolution from the perspective of flow dynamics~\cite{liu2022flow,lipman2022flow}. In a one-step flow update, the predicted velocity determines the \emph{direction} of restoration, while the update magnitude determines how far each latent location moves along that direction. Existing restoration pipelines often entangle these two factors by relying on a global condition and a uniform update strength. However, dark images require direction and magnitude to be handled differently. The restoration direction should be guided by semantic degradation intent and image-specific evidence, while the update magnitude should vary across space according to local degradation severity. This motivates a direction--magnitude decoupled formulation for super-resolution in the dark.
 
 In this paper, we propose \textbf{DeLang-SR}\footnote{Project page: \url{https://moriyaya.github.io/DeLang-SR/}.} (\textbf{De}gradation-\textbf{Lang}uage Steered \textbf{S}uper-\textbf{R}esolution), a language-steered one-step flow-matching framework. Rather than treating low-light corruption as arbitrary latent features or relying on manually written prompts, DeLang-SR converts measurable low-light statistics into structured degradation language. This produces a coarse restoration intent, discrete degradation semantics, and a continuous degradation descriptor. The language-derived condition is further complemented by visual degradation tokens extracted from the input image, providing instance-specific evidence about local degradation patterns. Together, these conditions steer the SD3~\cite{esser2024sd3} velocity field toward restoration-relevant directions.
 
 To control the magnitude of restoration, DeLang-SR predicts a degradation-aware update-strength map from the low-quality input and the continuous descriptor. This map acts as a locally varying Euler step-size field in the one-step flow update. As a result, heavily degraded shadow regions can receive larger corrective updates, whereas reliable regions are updated conservatively to preserve faithful details. The final restoration is therefore performed as a single direction-steered and magnitude-controlled flow update, avoiding iterative sampling and globally uniform restoration strength. This formulation provides both efficient inference and interpretable intermediate signals, including degradation-language conditions, visual tokens, and update-strength maps.
 
 Our contributions are summarized as follows:
 \begin{itemize}
 	\item We identify direction--magnitude entanglement as a key limitation of globally driven generative restoration for super-resolution in the dark, and formulate one-step restoration as a language-steered flow update with spatially varying update strength. 
 	
 	\item We propose \textbf{DeLang-SR}, which translates measurable low-light statistics into structured degradation language and combines language-derived conditions with visual degradation tokens to steer the restoration direction of a pretrained SD3 velocity field.
 	
 	\item We design a degradation-aware update-strength field that serves as a locally varying Euler step-size map, enabling magnitude-controlled restoration across severely degraded and relatively reliable regions.
 	
 	\item We conduct extensive paired and no-reference evaluations, controlled ablations, efficiency analysis, and diagnostic visualizations, demonstrating effective one-step super-resolution in the dark without oracle degradation annotations or iterative sampling.
 \end{itemize}

\section{Related Work}

\textbf{Dark and Real-World Image Super-Resolution.}
Image super-resolution has evolved from distortion-oriented reconstruction to real-world restoration, where the degradation process is unknown, spatially variant, and often entangled with noise, blur, compression, or illumination corruption. Early real-world SR methods improve robustness by synthesizing complex degradation pipelines or learning stronger restoration backbones~\cite{wang2021realesrgan,liang2021swinir,chen2023hat,zamir2022restormer}. Recent transformer-based SR models further improve long-range feature aggregation and high-frequency reconstruction~\cite{zhou2023srformer,chen2024recursive,zhang2024hit,long2025progressive,liu2025catanet}. In parallel, low-light enhancement methods focus on recovering visibility and illumination from underexposed observations~\cite{liu2021retinex,ma2022sci,wang2023ultra}. However, super-resolution \textit{in the dark} is not a simple combination of enhancement and upsampling: photometric degradation and structural detail loss are coupled, and cascaded pipelines often propagate enhancement artifacts into the SR stage. RELLISUR provides a paired benchmark for joint low-light enhancement and super-resolution~\cite{aakerberg2021rellisur}. Recent dark SR methods jointly model brightening and resolution amplification through task-specific architectures~\cite{yue2024unveiling,gao2024collaba}. Illumination-aware modulation and conditional-flow modeling provide further alternatives for handling the coupled degradation~\cite{gao2025dualstream,yue2024lolisrflow}. Hierarchical and nested optimization frameworks additionally coordinate the coupled task objectives or adapt their configurations~\cite{gao2024trico,gao2026snoc}. Nevertheless, most existing methods still learn a monolithic restoration mapping. They lack an explicit mechanism to determine how strongly different spatial regions should be restored under heterogeneous darkness.

\textbf{Diffusion- and Flow-Based Super-Resolution.}
Diffusion and flow models have turned large-scale generative priors into a practical foundation for perceptual image restoration.
Starting from denoising diffusion models~\cite{ho2020ddpm}, latent diffusion models~\cite{rombach2022ldm} improve high-resolution synthesis efficiency by operating in a compressed latent space, enabling large-scale text-to-image systems such as Stable Diffusion and SDXL~\cite{podell2023sdxl}. 
More recently, flow matching and rectified flow formulate generation through continuous velocity fields~\cite{lipman2022flow,liu2022flow}, and Stable Diffusion 3 scales this paradigm with rectified-flow transformers for high-resolution text-to-image synthesis~\cite{esser2024sd3}. 
These advances make pretrained diffusion and flow priors increasingly attractive for image restoration and super-resolution.

A growing body of SR and restoration works adapts such priors to real-world degradations. 
StableSR~\cite{wang2024stablesr} and DiffBIR~\cite{lin2024diffbir} exploit latent diffusion priors for real-world SR and blind restoration, ResShift~\cite{yue2023resshift} shortens the diffusion trajectory through residual shifting, and SeeSR~\cite{wu2024seesr} introduces degradation-aware semantic guidance for generative SR. 
To reduce the cost of iterative sampling, recent methods further compress diffusion restoration into few-step or one-step processes, including SinSR~\cite{wang2024sinsr}, TSD-SR~\cite{dong2025tsdsr}, S3Diff~\cite{zhang2024s3diff}, DOD~\cite{tang2026dod}, DHGM~\cite{li2026seeing}, and OMGSR~\cite{wu2025omgsr}. 
These methods improve perceptual realism and inference efficiency, but most still rely on globally organized sampling trajectories or shared update schedules. 
SARGD~\cite{zheng2024sargd} is a spatially adaptive diffusion approach that uses a binary artifact mask to guide latent replacement during iterative DDPM sampling. 
In contrast, DeLang-SR formulates heterogeneous dark SR as a direction--magnitude decoupled one-step flow update: degradation-language conditions steer the velocity direction, while a continuous degradation-conditioned field controls the local integration magnitude.

 \textbf{Language- and Degradation-Conditioned Restoration.}
 Language and degradation-aware representations have emerged as effective interfaces for controlling image restoration. 
 More broadly, adaptive hyper-guidance and constraint-learning formulations provide optimization tools for coupled image-enhancement problems~\cite{gao2023hyperguidance,liu2024constraint}.
 PromptIR~\cite{wang2023promptir} uses learnable prompts to adapt a blind restoration model to different corruptions, while DA-CLIP~\cite{luo2024daclip} exploits vision-language priors to obtain degradation-aware task guidance. 
 For generative SR and restoration, methods such as SeeSR~\cite{wu2024seesr}, PromptSR~\cite{chen2025promptsr}, S3Diff~\cite{zhang2024s3diff}, and DeFusion~\cite{luo2025defusion} further introduce semantic prompts, degradation descriptions, or visual instructions to better activate pretrained diffusion priors. 
 These works validate the value of language-level conditions, but their role is usually limited to conditioning the restoration model. 
 Such conditioning often specifies \textit{what} should be restored at a global level, yet it does not explicitly regulate \textit{how strongly} the generative prior should be applied across spatially varying regions. 
 DeLang-SR differs by assigning mathematically distinct roles to degradation-control signals: translated degradation language and visual tokens steer the SD3 velocity direction, while a degradation-aware spatial map controls the local Euler integration magnitude. 
 This direction-magnitude separation moves beyond prompt-conditioned restoration, embedding degradation language into the mechanics of the flow update itself.

\section{Method}

\begin{figure}[!t]
	\centering 
	\includegraphics[width=0.99\linewidth]{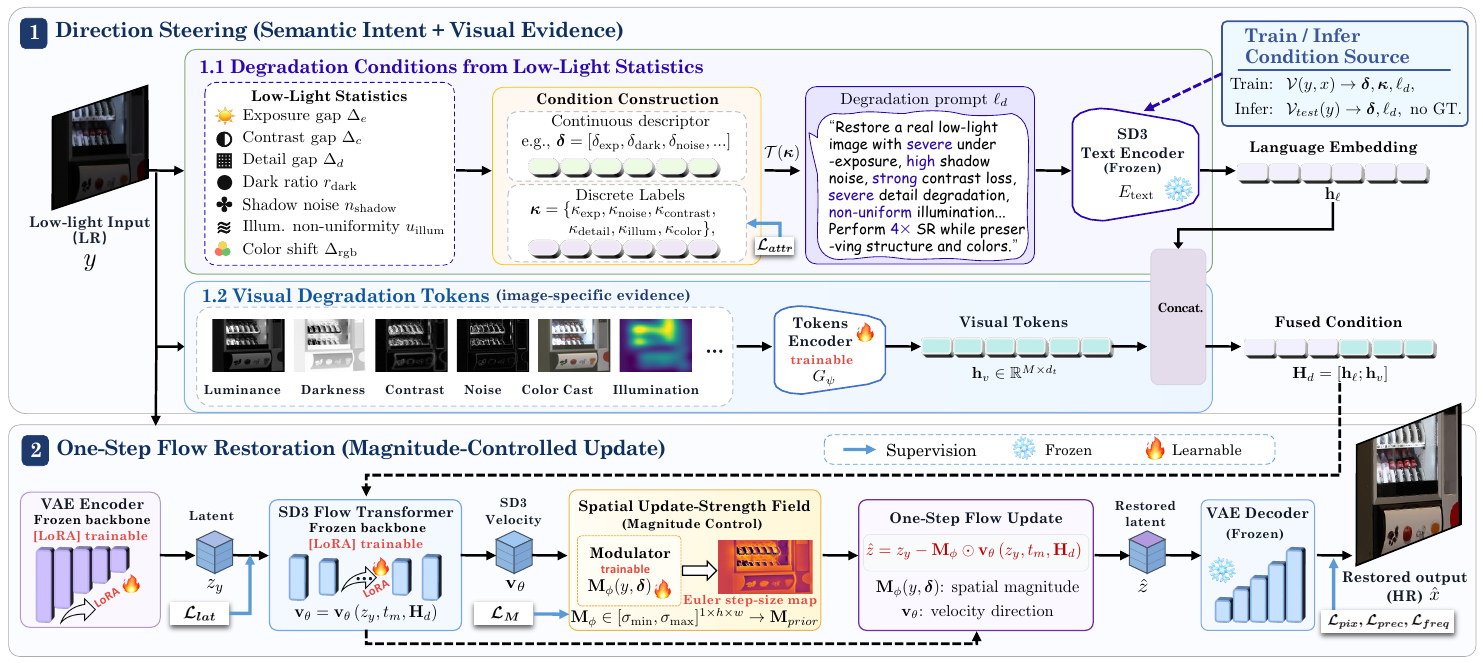}
	\vspace{-0.2cm} 
\caption{
	Pipeline of DeLang-SR.
	Degradation language and visual tokens form \(\mathbf{H}_d\) to steer the SD3 velocity direction, while the continuous descriptor and local evidence predict \(\mathbf{M}_{\phi}(y,\boldsymbol{\delta})\) to control spatial update magnitude.
	During inference, \((\boldsymbol{\delta},\ell_d)\) are estimated by \(\mathcal{V}_{test}(y)\) without GT, and the restored image is obtained by a single direction-steered and magnitude-controlled flow update.
}
	\label{fig:pipeline}
	\vspace{-0.6cm}
\end{figure}

\subsection{Overview}

Given a low-light low-resolution observation \(y\), DeLang-SR aims to recover a normally exposed high-resolution image \(\hat{x}\). Super-resolution in the dark is inherently spatially heterogeneous: deeply underexposed regions require stronger generative correction to recover visibility and suppressed structures, whereas relatively reliable regions should be updated conservatively to avoid over-correction and hallucinated textures. We therefore formulate the task as a \textit{degradation-conditioned one-step flow restoration} problem, where the restoration direction and spatial update magnitude are explicitly decoupled.

Let \(x\) denote the paired high-quality target during training. We write the SD3 VAE latents as \(z_y=\mathcal{E}_{vae}(y)\) and \(z_x=\mathcal{E}_{vae}(x)\). Under the SD3 flow-matching formulation~\cite{esser2024sd3,lipman2022flow,liu2022flow},  the transformer predicts a velocity field \(\mathbf{v}_{\theta}(z_t,t,\mathbf{H})\). Instead of iterative sampling, DeLang-SR performs a single degradation-controlled Euler update:
\begin{equation}
	\hat{z}
	=
	z_y
	-
	\mathbf{M}_{\phi}(y,\boldsymbol{\delta})
	\odot
	\mathbf{v}_{\theta}
	\left(
	z_y,t_m,\mathbf{H}_{d}
	\right),
	\qquad
	\hat{x}=\mathcal{D}_{vae}(\hat{z}),
	\label{eq:overview_update}
\end{equation}
where \(t_m\) is a fixed operating timestep, \(\boldsymbol{\delta}\) is a continuous degradation descriptor, \(\mathbf{H}_{d}\) is the fused degradation condition, and \(\mathbf{M}_{\phi}(y,\boldsymbol{\delta})\) is a spatial update-strength field. 
Here, \(\mathbf{H}_{d}\) steers the SD3 velocity direction, while \(\mathbf{M}_{\phi}\) controls the local integration magnitude. 
Figure~\ref{fig:pipeline} provides an overview of the degradation-condition construction, direction steering, magnitude control, and one-step flow update. 
Details of the one-step convention are provided in \appref{app:architecture}.

\subsection{Degradation Conditions from Low-Light Statistics}

A generic prompt such as ``restore a low-light image'' cannot distinguish underexposure, shadow noise, contrast collapse, non-uniform illumination, color cast, and detail degradation. Meanwhile, raw numerical statistics require an interface to interact with the language-conditioned prior of pretrained text-to-image models. We therefore convert measurable low-light statistics into structured degradation conditions.

For a paired sample \((y,x)\), we compute 
\(\mathbf{s}(y,x)=[\Delta_e,\Delta_c,\Delta_d,r_{\mathrm{dark}},n_{\mathrm{shadow}},u_{\mathrm{illum}},\Delta_{\mathrm{rgb}}]\), covering exposure, contrast, detail, darkness, shadow noise, illumination non-uniformity, and color shift. 
The statistics are mapped into structured degradation conditions:
{\small
\begin{equation}
	(\boldsymbol{\delta},\boldsymbol{\kappa},\ell_d)
	=
	\mathcal{V}(y,x),
	\quad
	\left\{
	\begin{aligned}
		\boldsymbol{\delta}
		&=
		[\delta_{\mathrm{exp}},\delta_{\mathrm{dark}},\delta_{\mathrm{noise}},
		\delta_{\mathrm{contrast}},\delta_{\mathrm{detail}},
		\delta_{\mathrm{illum}},\delta_{\mathrm{scale}}]\in\mathbb{R}^{7},\\
		\boldsymbol{\kappa}
		&=
		\{\kappa_{\mathrm{exp}},\kappa_{\mathrm{noise}},\kappa_{\mathrm{contrast}},
		\kappa_{\mathrm{detail}},\kappa_{\mathrm{illum}},\kappa_{\mathrm{color}}\},
		\quad
		\ell_d=\mathcal{T}(\boldsymbol{\kappa}).
	\end{aligned}
	\right.
	\label{eq:condition_construction}
\end{equation}
}
Here \(\boldsymbol{\delta}\) is a continuous descriptor that guides the spatial update-strength field, \(\boldsymbol{\kappa}\) denotes discrete degradation labels for prompt generation and attribute alignment, and \(\ell_d\) is the resulting degradation-language prompt. 
Color cast is represented as a discrete label rather than a continuous entry of \(\boldsymbol{\delta}\).
 The discrete labels \(\boldsymbol{\kappa}\) are obtained by quantizing the corresponding statistics and inserted into a deterministic template \(\ell_d=\mathcal{T}(\boldsymbol{\kappa})\). 
For example, a generated degradation-language prompt is:
\begin{center}
	\fcolorbox{blue!35}{blue!3}{
		\begin{minipage}{0.92\linewidth}
			\small
			\textbf{Prompt.}
			Restore a real low-light image with 
			\promptslot{severe} underexposure, 
			\promptslot{high} shadow noise, 
			\promptslot{strong} contrast loss, 
			\promptslot{severe} detail degradation, 
			\promptslot{non-uniform} illumination, 
			and a \promptslot{warm} color cast. 
			Perform \promptslot{\(4\times\)} super-resolution while preserving scene structure and natural colors.
		\end{minipage}
	}
\end{center}
The highlighted slots are automatically instantiated from the discretized degradation labels \(\boldsymbol{\kappa}\), which are derived from measurable low-light statistics rather than manual annotations. 
Detailed degradation statistics, descriptor-label correspondences, and prompt construction rules are provided in \appref{app:deg_language}, with the main definitions summarized in Tables~\ref{tab:app_degradation_statistics} and~\ref{tab:app_descriptor_labels}.

At inference time, the high-quality reference \(x\) is unavailable. We estimate degradation conditions from the low-quality input itself, i.e., \((\boldsymbol{\delta},\ell_d)=\mathcal{V}_{test}(y)\). Since LQ-only estimation is approximate, \(\ell_d\) is treated as a coarse restoration intent rather than an oracle annotation. Thus, DeLang-SR requires no manual prompt, oracle degradation label, or high-quality reference during inference.

\subsection{Direction Steering and Magnitude Control}
\textbf{Language and Visual Evidence for Direction Steering.}
The degradation prompt \(\ell_d\) provides a global restoration intent, but language alone cannot localize degradation patterns in the current image.
We therefore introduce visual degradation tokens to provide image-specific evidence.
The prompt is encoded by the frozen SD3 text encoder as
\(\mathbf{h}_{\ell}=E_{\mathrm{text}}(\ell_d)\), while a visual token encoder
\(G_{\psi}\) produces \(M\) learnable tokens
\(\mathbf{h}_{v}=G_{\psi}(y)\in\mathbb{R}^{M\times d_t}\)
from the RGB input, luminance, darkness, local contrast, and global low-light statistics. 
The detailed architecture of \(G_{\psi}\) is provided in \appref{app:vdt}.
The language embedding and visual tokens are concatenated as:
\begin{equation}
	\mathbf{H}_{d}
	=
	[
	\mathbf{h}_{\ell};
	\mathbf{h}_{v}
	].
	\label{eq:fused_condition}
\end{equation}
Here \(\mathbf{h}_{\ell}\) specifies the coarse semantic intent of what degradations should be corrected, while \(\mathbf{h}_{v}\) grounds this intent with image-specific evidence about where such degradations appear. The fused condition \(\mathbf{H}_{d}\) is injected into the SD3 flow transformer to steer the restoration direction.

\noindent\textbf{Spatial Update-Strength Field for Magnitude Control.}
While \(\mathbf{H}_{d}\) controls the restoration direction, one-step flow restoration also requires controlling how strongly the velocity is applied at each latent location. We therefore predict a spatial update-strength field:
\begin{equation}
	\mathbf{M}_{\phi}(y,\boldsymbol{\delta})
	=
	\sigma_{\min}
	+
	(\sigma_{\max}-\sigma_{\min})
	\cdot
	\mathrm{Sigmoid}
	\left(
	g_{\phi}
	\left(
	E_{\phi}(y),
	P_{\phi}(\boldsymbol{\delta})
	\right)
	\right).
	\label{eq:sigma_prediction}
\end{equation}
Here \(E_{\phi}\) and \(P_{\phi}\) denote the visual evidence extractor and descriptor-conditioning branch, respectively; their FiLM-based implementation is detailed in \appref{app:sigma_modulator}.The resulting field \(\mathbf{M}_{\phi}(y,\boldsymbol{\delta})\in[\sigma_{\min},\sigma_{\max}]^{1\times h\times w}\) is predicted at the latent resolution and broadcast along latent channels.

From the one-step flow perspective, \(\mathbf{M}_{\phi}\) acts as a spatially varying Euler step-size field. It does not alter the predicted direction; instead, it controls the local integration magnitude. Thus, the pretrained prior is not applied through a uniform global update, but locally scaled according to degradation severity and visual evidence. Implementation details are provided in \appref{app:sigma_modulator}.

\subsection{One-Step Flow Restoration Dynamics}

Given \(z_y\), \(\mathbf{H}_{d}\), and \(\mathbf{M}_{\phi}(y,\boldsymbol{\delta})\), DeLang-SR predicts the SD3 velocity at a fixed operating timestep \(t_m\):
\begin{equation}
	\mathbf{v}_{\theta}
	=
	\mathbf{v}_{\theta}
	\left(
	z_y,t_m,\mathbf{H}_{d}
	\right).
	\label{eq:predicted_flow}
\end{equation}
The restored latent is then obtained by the one-step update in Eq.~\eqref{eq:overview_update}. Equivalently, \(\mathbf{v}_{\theta}(z_y,t_m,\mathbf{H}_{d})\) determines the restoration direction in the pretrained flow space, while \(\mathbf{M}_{\phi}(y,\boldsymbol{\delta})\) determines how far each latent location moves along that direction.
The operating timestep \(t_m\) provides a fixed point on the pretrained SD3 flow trajectory. DeLang-SR does not use \(t_m\) to encode image-specific degradation; instead, region-dependent restoration strength is controlled by \(\mathbf{M}_{\phi}\). For parameter-efficient adaptation, we freeze the SD3 text encoder and VAE decoder, insert LoRA adapters~\cite{hu2022lora} into the VAE encoder and SD3 flow transformer, and train the degradation-control modules from scratch.

\noindent\textbf{Learning Objective.}
To train the direction-steered and magnitude-controlled update, we supervise three complementary aspects: latent flow alignment, image reconstruction, and degradation-control regularization. 
For each paired sample \((y,x)\), we construct \((\boldsymbol{\delta},\boldsymbol{\kappa},\ell_d)=\mathcal{V}(y,x)\), encode \(z_x=\mathcal{E}_{fix}(x)\) and \(z_y=\mathcal{E}_{adapt}(y)\), and build a spatially degradation-dependent flow state. Here, \(\mathcal{E}_{fix}\) denotes the frozen pretrained VAE encoder used to obtain target latents, and \(\mathcal{E}_{adapt}\) denotes its LoRA-adapted counterpart for low-light inputs:
\begin{equation}
	z_{\mathbf{M}}
	=
	(1-\mathbf{M}_{\phi}(y,\boldsymbol{\delta}))\odot z_x
	+
	\mathbf{M}_{\phi}(y,\boldsymbol{\delta})\odot \epsilon,
	\qquad
	\epsilon\sim\mathcal{N}(0,I),
\end{equation}
with \(\mathcal{L}_{lat}=\|z_y-z_{\mathbf{M}}\|_2^2\).
The restored image is supervised by \(\mathcal{L}_{pix}=\|\hat{x}-x\|_1\), a DISTS-style perceptual loss \(\mathcal{L}_{perc}=\mathrm{DISTS}_{DINOv3}(\hat{x},x)\)~\cite{ding2020image,simeoni2025dinov3}, and a frequency-decoupled loss: 
\begin{equation}
	\mathcal{L}_{freq}
	=
	\|
	\mathrm{Low}(\hat{x})-\mathrm{Low}(x)
	\|_1
	+
	\|
	\mathrm{High}(\hat{x})-\mathrm{High}(x)
	\|_1.
\end{equation}
We further regularize the update-strength field with \(\mathcal{L}_{M}=\|\mathbf{M}_{\phi}(y,\boldsymbol{\delta})-\mathbf{M}_{prior}(y)\|_1\), where \(\mathbf{M}_{prior}(y)=\sigma_{\min}+(\sigma_{\max}-\sigma_{\min})d_y\). 
Finally, an attribute head predicts the discrete labels from the visual tokens, yielding \(\mathcal{L}_{attr}=\sum_{k=1}^{K}\mathrm{CE}(\hat{\kappa}^{k},\kappa^{k})\), where \(K=6\) and \(\mathrm{CE}\) denotes cross-entropy.
The total objective is:
\begin{equation}
	\mathcal{L}
	=
	\lambda_{lat}\mathcal{L}_{lat}
	+
	\lambda_{pix}\mathcal{L}_{pix}
	+
	\lambda_{freq}\mathcal{L}_{freq}
	+
	\lambda_{perc}\mathcal{L}_{perc}
	+
	\lambda_{M}\mathcal{L}_{M}
	+
	\lambda_{attr}\mathcal{L}_{attr}. \label{eq:total_objective}
\end{equation} 
At inference time, DeLang-SR estimates \((\boldsymbol{\delta},\ell_d)\) from \(y\), constructs \(\mathbf{H}_d\), \(\mathbf{M}_\phi\), and performs one-step update in Eq.~\eqref{eq:overview_update} without high-quality references, oracle labels, manual prompts, or iterative sampling.

\section{Experiments}

\subsection{Experimental Protocol}

\noindent\textbf{Datasets.}
We evaluate DeLang-SR on both paired and real dark super-resolution settings. 
For paired evaluation, we use RELLISUR~\cite{aakerberg2021rellisur}, which contains 3,610 training pairs and 425 testing pairs under both \(\times2\) and \(\times4\) settings. 
The low-light image is used as the degraded observation \(y\), and the paired normal-light high-resolution image is used as the target \(x\). 
To assess generalization to real nighttime scenes, we further evaluate on 100 images from DarkFace~\cite{poor_visibility_benchmark} without reference.

\noindent\textbf{Metrics.}
For paired evaluation, we report PSNR, SSIM~\cite{wang2004image}, LPIPS~\cite{zhang2018perceptual}, and DISTS~\cite{ding2020image} to measure fidelity, structure, and perceptual consistency. 
Since dark SR also requires natural illumination and chromatic fidelity, we additionally report exposure error (ExpErr) and color consistency error (ColorErr), computed against the paired high-quality target. 
For no-reference evaluation, we report NIQE~\cite{mittal2012making}, MUSIQ~\cite{ke2021musiq}, MANIQA~\cite{yang2022maniqa}, and CLIP-IQA~\cite{wang2023clipiqa} on RELLISUR outputs and real DarkFace images. 
We also report runtime and flow-update steps to assess inference efficiency.

\noindent\textbf{Baselines.}
We compare DeLang-SR with three groups of baselines: 
(i) general SR/restoration backbones, including Restormer~\cite{zamir2022restormer}, SwinIR~\cite{liang2021swinir}, SRFormer~\cite{zhou2023srformer}, HAT~\cite{chen2023hat}, RGT~\cite{chen2024recursive}, HiT-SR~\cite{zhang2024hit}, PFT-SR~\cite{long2025progressive}, ASID~\cite{park2025efficient}, and CATANet~\cite{liu2025catanet}; 
(ii) cascaded LLE+SR pipelines built from RUAS~\cite{liu2021retinex}, SCI~\cite{ma2022sci}, and LLFormer~\cite{wang2023ultra} with strong SR backbones; 
and (iii) recent diffusion- or degradation-guided restoration/SR methods, including S3Diff~\cite{zhang2024s3diff},  DOD~\cite{tang2026dod}, and DHGM~\cite{li2026seeing}. 
All methods are evaluated on the same test images and output resolutions.

\noindent\textbf{Implementation.}
DeLang-SR is initialized from a pretrained Stable Diffusion 3 (SD3) model~\cite{esser2024sd3}. 
We freeze the SD3 text encoder and VAE decoder, and insert lightweight LoRA adapters~\cite{hu2022lora} into the VAE encoder and SD3 flow transformer. 
The newly introduced degradation-control modules, including the visual token encoder \(G_\psi\), spatial update-strength predictor \(\mathbf{M}_{\phi}\), and attribute prediction head, are optimized jointly with the LoRA adapters. 
We use \(M=4\) visual degradation tokens and a 7-dimensional descriptor \(\boldsymbol{\delta}\). 
At inference time, DeLang-SR estimates \((\boldsymbol{\delta},\ell_d)=\mathcal{V}_{test}(y)\) from the input \(y\) and restores the image with a single direction-steered and magnitude-controlled SD3 flow update. 
Unless otherwise specified, main comparisons use the best full checkpoint, while controlled ablations are conducted on RELLISUR \(\times2\) under the same 5K-step training budget.

\subsection{Diagnostic Analysis of Degradation-Control Mechanisms}

\begin{figure}[!t]
 	\centering 
 	\includegraphics[width=0.98\linewidth]{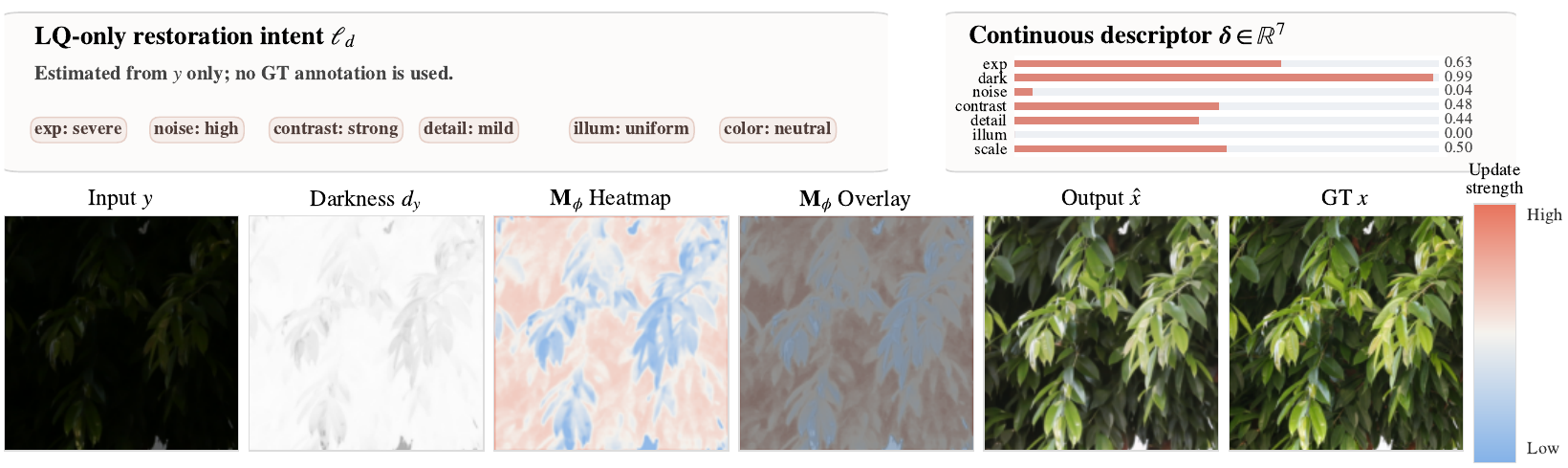}
 	\vspace{-0.2cm}
\caption{
	Visualization of the update-strength map \(\mathbf{M}_{\phi}(y,\boldsymbol{\delta})\).
	Both \(\ell_d\) and \(\boldsymbol{\delta}\) are estimated from \(y\) without GT.
	The map provides an interpretable spatial step-size field for one-step SD3 restoration.
}
 	\label{fig:modulation_map}
 	\vspace{-0.2cm}
 \end{figure}

\noindent\textbf{Spatial update-strength field.}
Before comparing with competing methods, we first visualize the proposed spatial update-strength field \(\mathbf{M}_{\phi}(y,\boldsymbol{\delta})\), which is the key component for magnitude control in DeLang-SR. 
As shown in Fig.~\ref{fig:modulation_map}, we present the input image, darkness map, predicted update-strength field, overlay visualization, restored output, and ground truth. 
The predicted field is spatially non-uniform and provides an interpretable local step-size map for one-step SD3 restoration. 
This supports our direction--magnitude decoupling view: the fused condition steers the restoration direction, while \(\mathbf{M}_{\phi}\) controls the local update magnitude according to degradation severity. 
Additional update-strength visualizations are provided in \appref{app:update_strength_visual}.

\begin{figure}[!t]
 	\centering
 	\vspace{-0.1cm}
 	\includegraphics[width=0.98\linewidth]{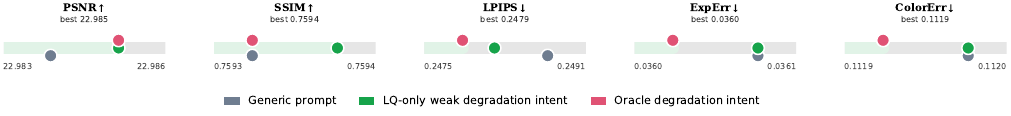}
 	\vspace{-0.2cm}
 	\caption{
 		Inference-time condition source analysis. 
 		The close overlap among generic, LQ-only, and oracle intents shows that DeLang-SR does not require oracle degradation annotations.
 	}
 	\label{fig:condition_source_capsules}
 	\vspace{-0.2cm}
 \end{figure}

\noindent\textbf{LQ-only condition construction.}
Since oracle degradation statistics are unavailable at test time, we evaluate three condition sources with the same checkpoint: a generic prompt, the proposed LQ-only degradation intent, and an oracle intent constructed from paired data. 
As shown in Fig.~\ref{fig:condition_source_capsules}, their close overlap indicates that DeLang-SR does not rely on high-quality references or oracle annotations, and the generated degradation language serves as a coarse restoration intent rather than a precise label.

\begin{table}[!t]
	\centering
	\setlength{\tabcolsep}{2.0pt}
	\renewcommand{\arraystretch}{0.92}
	\vspace{-0.1cm}
		\caption{Paired quantitative comparison on \textbf{\textit{RELLISUR}}. 
			\colorbox{rankone}{\textbf{Pink}}, \colorbox{ranktwo}{\textbf{orange}}, and \colorbox{rankthree}{\textbf{yellow}} mark the top three results. 
			\(\clubsuit\), \(\blacktriangle\), and \(\blacklozenge\) denote cascaded LLE+SR, general SR, and diffusion/degradation-based methods, respectively. 
			\(\bm{^{\dagger}}\)/\(\bm{^{\ddagger}}\) indicate LLE-trained/normal-light SR-trained models.}
	\label{tab:rellisur_main}
	\vspace{-0.2cm}
	\resizebox{0.99\linewidth}{!}{%
		\begin{tabular}{@{}lc cccccc cccccc@{}}
			\toprule[1.1pt]
			\multirow{2}{*}{\textbf{\emph{Method}}}
			& \multirow{2}{*}{\textbf{Type}}
			& \multicolumn{6}{c}{\textbf{\emph{RELLISUR @ \(\times2\)}}}
			& \multicolumn{6}{c}{\textbf{\emph{RELLISUR @ \(\times4\)}}} \\
			\cmidrule(lr){3-8} \cmidrule(lr){9-14}
			&
			& \textbf{PSNR$\uparrow$}
			& \textbf{SSIM$\uparrow$}
			& \textbf{LPIPS$\downarrow$}
			& \textbf{DISTS$\downarrow$}
			& \textbf{ExpErr$\downarrow$}
			& \textbf{ColorErr$\downarrow$}
			& \textbf{PSNR$\uparrow$}
			& \textbf{SSIM$\uparrow$}
			& \textbf{LPIPS$\downarrow$}
			& \textbf{DISTS$\downarrow$}
			& \textbf{ExpErr$\downarrow$}
			& \textbf{ColorErr$\downarrow$} \\
			\midrule
			
\(\bm{^{\dagger}}\)RUAS\(\rightarrow\)\(\bm{^{\ddagger}}\)SwinIR & \castype & 17.160 & 0.6170 & 0.5811 & 0.3048 & 0.1024 & 0.3074 & 17.058 & 0.6624 & 0.6633 & 0.3756 & 0.0871 & 0.2611 \\
\(\bm{^{\dagger}}\)SCI\(\rightarrow\)\(\bm{^{\ddagger}}\)HAT & \castype & 14.963 & 0.4389 & 0.5968 & 0.3381 & 0.1644 & 0.5038 & 14.750 & 0.4523 & 0.7040 & 0.3591 & 0.1644 & 0.5057 \\
\(\bm{^{\dagger}}\)LLFormer\(\rightarrow\)\(\bm{^{\ddagger}}\)HAT & \castype & \colorbox{rankthree}{\textbf{21.221}} & 0.7205 & 0.4617 & 0.2660 & \colorbox{rankthree}{\textbf{0.0549}} & \colorbox{rankthree}{\textbf{0.1683}} & 20.123 & 0.7182 & 0.5847 & 0.3639 & 0.0636 & \colorbox{rankthree}{\textbf{0.1951}} \\
\midrule
SwinIR (ICCVW'21) & \srtype & 18.389 & 0.6407 & 0.5866 & 0.2862 & 0.0666 & 0.2049 & 17.513 & 0.6637 & 0.6968 & 0.3922 & 0.0797 & 0.2445 \\
Restormer (CVPR'22) & \srtype & 21.219 & \colorbox{rankthree}{\textbf{0.7279}} & 0.3901 & \colorbox{rankthree}{\textbf{0.2441}} & 0.0593 & 0.1800 & \colorbox{ranktwo}{\textbf{20.276}} & \colorbox{rankthree}{\textbf{0.7207}} & \colorbox{rankthree}{\textbf{0.5029}} & \colorbox{rankthree}{\textbf{0.3179}} & 0.0660 & 0.2025 \\
SRFormer (ICCV'23) & \srtype & 19.559 & 0.7049 & 0.4724 & 0.2564 & 0.0565 & 0.1735 & 18.708 & 0.7055 & 0.6228 & 0.3384 & \colorbox{ranktwo}{\textbf{0.0630}} & 0.1953 \\
HAT (CVPR'23) & \srtype & 20.217 & 0.7193 & 0.4598 & 0.2707 & 0.0574 & 0.1753 & 19.734 & 0.7154 & 0.5730 & 0.3567 & 0.0661 & 0.2021 \\
RGT (ICLR'24) & \srtype & 18.599 & 0.6702 & 0.5479 & 0.2861 & 0.0640 & 0.1957 & 19.154 & 0.7163 & 0.5464 & 0.3389 & 0.0669 & 0.2016 \\
HiT-SR (ECCV'24) & \srtype & 16.291 & 0.6242 & 0.5903 & 0.4014 & 0.0755 & 0.2311 & 17.647 & 0.6826 & 0.6432 & 0.3767 & 0.0705 & 0.2218 \\
ASID (AAAI'25) & \srtype & 19.045 & 0.6971 & 0.4470 & 0.2528 & 0.0564 & 0.1723 & 18.155 & 0.7002 & 0.5634 & 0.3309 & \colorbox{rankthree}{\textbf{0.0634}} & \colorbox{ranktwo}{\textbf{0.1943}} \\
PFT-SR (CVPR'25) & \srtype & 17.901 & 0.6493 & 0.5491 & 0.2841 & 0.0703 & 0.2151 & 18.534 & 0.7130 & 0.5577 & 0.3579 & 0.0671 & 0.2034 \\
CATANet (CVPR'25) & \srtype & 19.812 & 0.7108 & 0.4216 & 0.2559 & 0.0612 & 0.1866 & 18.892 & 0.7075 & 0.5505 & 0.3459 & 0.0679 & 0.2075 \\
\midrule
DHGM (AAAI'26) & \difftype & \colorbox{ranktwo}{\textbf{22.663}} & \colorbox{ranktwo}{\textbf{0.7625}} & \colorbox{rankthree}{\textbf{0.3726}} & 0.2484 & \colorbox{ranktwo}{\textbf{0.0437}} & \colorbox{ranktwo}{\textbf{0.1368}} & \colorbox{rankthree}{\textbf{20.146}} & \colorbox{rankone}{\textbf{0.7328}} & 0.5443 & 0.3505 & 0.0654 & 0.1999 \\
DOD (AAAI'26) & \difftype & 20.806 & 0.7199 & \colorbox{ranktwo}{\textbf{0.2557}} & \colorbox{ranktwo}{\textbf{0.1570}} & 0.0665 & 0.2000 & 17.844 & 0.6218 & \colorbox{ranktwo}{\textbf{0.4880}} & \colorbox{ranktwo}{\textbf{0.2705}} & 0.0734 & 0.2235 \\
\midrule
\textbf{DeLang-SR (Ours)} & \difftype & \colorbox{rankone}{\textbf{23.865}} & \colorbox{rankone}{\textbf{0.7700}} & \colorbox{rankone}{\textbf{0.2186}} & \colorbox{rankone}{\textbf{0.1462}} & \colorbox{rankone}{\textbf{0.0268}} & \colorbox{rankone}{\textbf{0.0856}} & \colorbox{rankone}{\textbf{21.259}} & \colorbox{ranktwo}{\textbf{0.7242}} & \colorbox{rankone}{\textbf{0.3290}} & \colorbox{rankone}{\textbf{0.2121}} & \colorbox{rankone}{\textbf{0.0441}} & \colorbox{rankone}{\textbf{0.1361}} \\

			\bottomrule[1.1pt]
		\end{tabular}
	}
	\vspace{-0.2cm}
\end{table}

\subsection{Main Comparison on Super-Resolution in the Dark}

\begin{figure}[!t]
	\centering
	\vspace{-0.1cm}
	\includegraphics[width=0.99\linewidth]{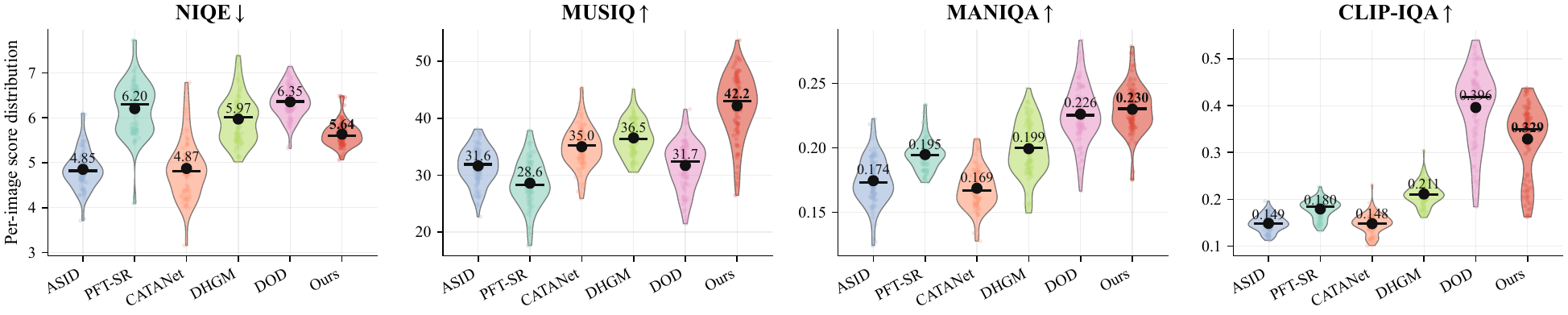}
	\vspace{-0.2cm}
	\caption{
		Distribution of no-reference perceptual quality scores on DarkFace under the \(\times2\) setting. }
	\label{fig:nr_distribution}
	\vspace{-0.2cm}
\end{figure}

We then evaluate the overall restoration performance of DeLang-SR. 
Table~\ref{tab:rellisur_main} reports paired quantitative results on RELLISUR under \(\times2\) and \(\times4\) settings, where fidelity, perceptual quality, illumination naturalness, and color consistency are jointly evaluated. 
Table~\ref{tab:nr_quality} reports no-reference perceptual quality on RELLISUR, Figure~\ref{fig:nr_distribution} shows the distribution of no-reference scores on real DarkFace images, and Table~\ref{tab:efficiency} provides the computational profile under the same output resolution.  As an SD3-based method, DeLang-SR has a larger computational footprint than lightweight SR backbones, but it avoids iterative sampling and adapts the pretrained prior with LoRA and lightweight control modules. 

\begin{figure}[!t]
	\centering
	\includegraphics[width=\linewidth,trim=1 2 1 5, clip]{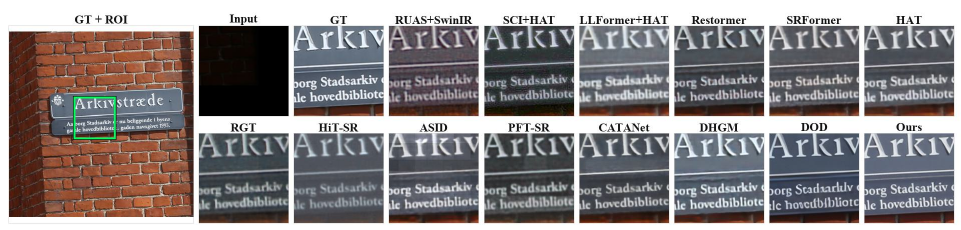}
	\includegraphics[width=\linewidth,trim=1 2 1 5, clip]{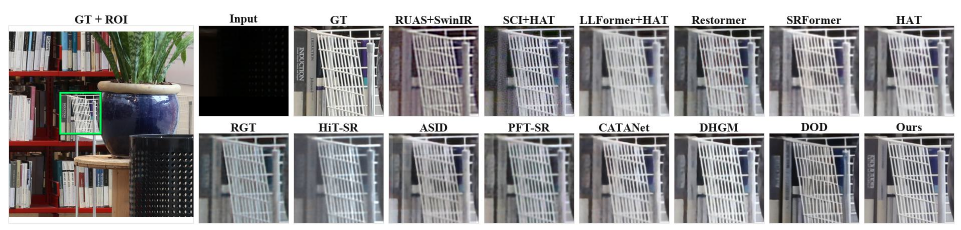}
			\includegraphics[width=\linewidth,trim=1 2 1 5, clip]{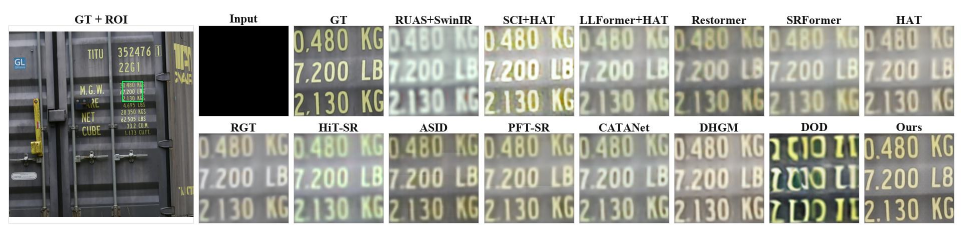}
		\includegraphics[width=\linewidth,trim=1 2 1 5, clip]{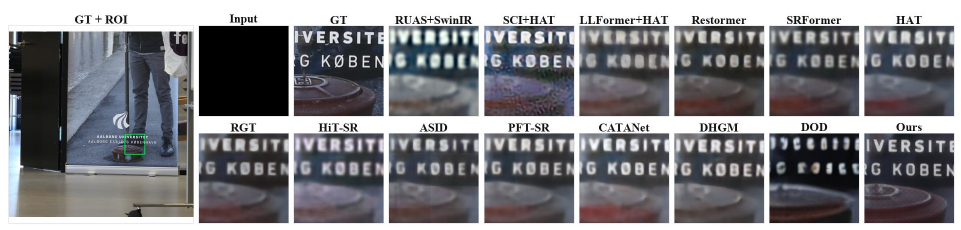}
	\vspace{-0.5cm}
\caption{
	Visual comparison of all competing methods on {{RELLISUR}} (Top two:×2, Bottom two:×4). More RELLISUR visual comparisons under \(\times2\) and \(\times4\) settings are provided in \appref{app:rellisur_visual}.
}
	\label{fig:rellisur_x2_roi_comparison}
	\vspace{-0.2cm}
\end{figure}

\begin{figure}[!t]
 	\centering
 	 	\includegraphics[width=\linewidth,trim=1 2 1 5, clip]{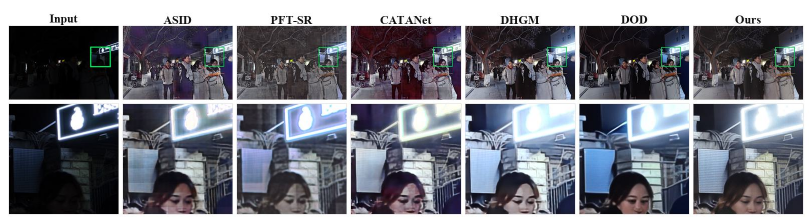} 
 	\vspace{-0.6cm}
 	\caption{
		Visual comparison on \textbf{\textit{DarkFace}} under the \(\times2\) setting.
 	}
 	\label{fig:darkface_x2_roi_comparison}
 	\vspace{-0.2cm}
 \end{figure}

More RELLISUR visual comparisons under both \(\times2\) and \(\times4\) settings are provided in \appref{app:rellisur_visual}, with examples shown in Figs.~\ref{fig:app_rellisur_x2_visual} and~\ref{fig:app_rellisur_x4_visual}.
Compared with diffusion- or degradation-guided baselines, DeLang-SR performs restoration through a single direction-steered and magnitude-controlled SD3 flow update, achieving a stronger balance between fidelity, perceptual quality, exposure recovery, and color consistency. 
Figures~\ref{fig:rellisur_x2_roi_comparison} and~\ref{fig:darkface_x2_roi_comparison} provide qualitative comparisons on RELLISUR and DarkFace, respectively. 
DeLang-SR preserves sharper local structures and more faithful contrast under severe dark degradation, and generalizes well to real nighttime scenes.

\begin{table}[!t]
	\centering
	\setlength{\tabcolsep}{3.2pt}
	\renewcommand{\arraystretch}{0.62}
	\vspace{-0.1cm}
	\caption{No-reference perceptual quality comparison. 
		Top-three results are highlighted. }
	\label{tab:nr_quality}
	\vspace{-0.2cm}
	\resizebox{0.98\linewidth}{!}{%
		\begin{tabular}{@{}lc cccc cccc@{}}
			\toprule[1.1pt]
			\multirow{2}{*}{\textbf{\emph{Method}}}
			& \multirow{2}{*}{\textbf{Type}}
			& \multicolumn{4}{c}{\textbf{\emph{RELLISUR @ \(\times2\)}}}
			& \multicolumn{4}{c}{\textbf{\emph{RELLISUR @ \(\times4\)}}} \\
			\cmidrule(lr){3-6} \cmidrule(lr){7-10}
			&
			& \textbf{NIQE$\downarrow$}
			& \textbf{MUSIQ$\uparrow$}
			& \textbf{MANIQA$\uparrow$}
			& \textbf{CLIP-IQA$\uparrow$}
			& \textbf{NIQE$\downarrow$}
			& \textbf{MUSIQ$\uparrow$}
			& \textbf{MANIQA$\uparrow$}
			& \textbf{CLIP-IQA$\uparrow$} \\
			\midrule
			
HiT-SR (ECCV'24) & \srtype & \colorbox{rankthree}{\textbf{7.527}} & 26.2128 & \colorbox{rankthree}{\textbf{0.2743}} & \colorbox{rankthree}{\textbf{0.2580}} & 9.391 & 21.2099 & 0.2231 & 0.1851 \\
PFT-SR (CVPR'25) & \srtype & 8.298 & 31.3606 & 0.2217 & 0.2283 & 9.980 & 23.3443 & \colorbox{rankthree}{\textbf{0.2460}} & \colorbox{rankthree}{\textbf{0.2844}} \\
CATANet (CVPR'25) & \srtype & 8.015 & 35.5178 & 0.2003 & 0.1891 & \colorbox{rankthree}{\textbf{8.735}} & 21.5868 & 0.2094 & 0.1966 \\
\midrule
DHGM (AAAI'26) & \difftype & 8.795 & \colorbox{rankthree}{\textbf{41.7819}} & 0.2140 & 0.2050 & 9.929 & \colorbox{rankthree}{\textbf{25.3630}} & 0.2297 & 0.2277 \\
DOD (AAAI'26) & \difftype & \colorbox{ranktwo}{\textbf{7.357}} & \colorbox{ranktwo}{\textbf{50.2245}} & \colorbox{ranktwo}{\textbf{0.2819}} & \colorbox{rankone}{\textbf{0.4224}} & \colorbox{ranktwo}{\textbf{8.610}} & \colorbox{ranktwo}{\textbf{29.0542}} & \colorbox{ranktwo}{\textbf{0.2589}} & \colorbox{rankone}{\textbf{0.4935}} \\
\midrule
\textbf{DeLang-SR (Ours)} & \difftype & \colorbox{rankone}{\textbf{7.285}} & \colorbox{rankone}{\textbf{54.7134}} & \colorbox{rankone}{\textbf{0.2967}} & \colorbox{ranktwo}{\textbf{0.4188}} & \colorbox{rankone}{\textbf{7.113}} & \colorbox{rankone}{\textbf{40.9621}} & \colorbox{rankone}{\textbf{0.2800}} & \colorbox{ranktwo}{\textbf{0.4253}} \\

			\bottomrule[1.1pt]
		\end{tabular}
	}
	\vspace{-0.2cm}
\end{table}

\begin{table}[!t]
	\centering
	\setlength{\tabcolsep}{3.6pt}
	\renewcommand{\arraystretch}{0.94}
	\vspace{-0.1cm}
	\caption{Computational profile under the same output resolution. 
		Trainable parameters exclude frozen pretrained priors where applicable. }
	\label{tab:efficiency}
	\vspace{-0.2cm}
	\resizebox{0.98\linewidth}{!}{%
		\begin{tabular}{@{}lcccccccc@{}}
			\toprule[1.1pt]
			\textbf{\emph{Metric}}
			& \textbf{HiT-SR}
			& \textbf{ASID}
			& \textbf{PFT-SR}
			& \textbf{CATANet}
			& \textbf{RGT}
			& \textbf{DHGM}
			& \textbf{DOD}
			& \textbf{DeLang-SR} \\
			\midrule
			
			\textbf{Trainable Params. (M)\(\downarrow\)}
			& 0.772 & 0.298 & 19.624 & 0.477 & 10.051 & 10.75 & 958.684 & 4.48 \\
			
			\textbf{FLOPs (G)\(\downarrow\)}
			& 18.290 & 3.965 & 106.359 & 21.266 & 77.919 & 87.78 & 826.185 & 20390.77 \\
			
			\textbf{Inference Time (s)\(\downarrow\)}
			& 0.044 & 0.007 & 0.362 & 0.042 & 0.615 & 0.062 & 0.088 & 0.448 \\
			
			\bottomrule[1.1pt]
		\end{tabular}
	}
	\vspace{-0.2cm}
\end{table}

\FloatBarrier
\subsection{Component Analysis and Diagnostic Ablation}

We analyze different degradation-control components under a controlled 5K-step training budget on RELLISUR \(\times2\). 
Table~\ref{tab:component_ablation} includes a base model implemented from the S3Diff codebase~\cite{zhang2024s3diff}. 
Full DeLang-SR clearly improves over this base model across fidelity, perceptual, exposure, and color metrics.
%
Removing degradation language, visual tokens, or attribute alignment weakens the overall balance, confirming the complementary roles of language intent, image-specific evidence, and semantic token supervision. 
Because the attribute head operates on \(\mathbf{h}_v\), the variant without visual tokens necessarily disables attribute alignment as well.
Removing descriptor conditioning from \(\mathbf{M}_{\phi}\) improves exposure/color errors slightly but reduces fidelity and perceptual quality, indicating its role in balancing conservative correction and structure preservation.
Overall, the full model achieves the best trade-off across multiple metrics. 

\begin{figure}[!t]
 	\centering
	\includegraphics[width=0.98\linewidth,trim=1 2 1 8, clip]{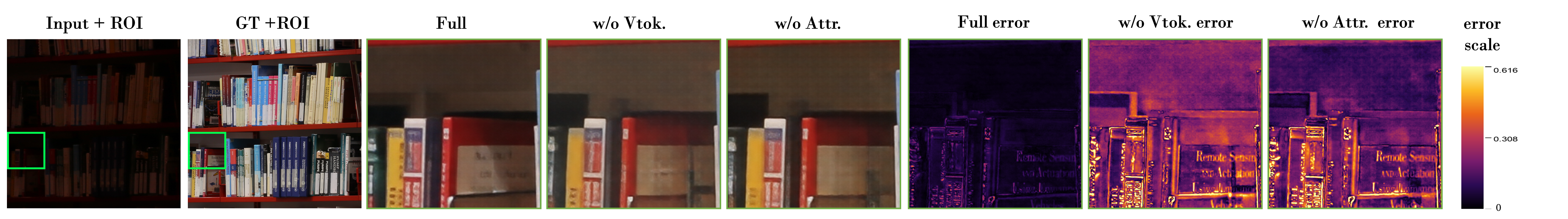}
	\vspace{-0.15cm}
\caption{
	Qualitative ablation among different cases in Table~\ref{tab:component_ablation}.
ROI crops and error maps show that visual tokens and attribute alignment improve local texture recovery.
}
 	\label{fig:ablation_roi_error}
 	\vspace{-0.15cm}
 \end{figure}

\begin{figure}[!t]
	\centering
	\includegraphics[width=0.98\linewidth]{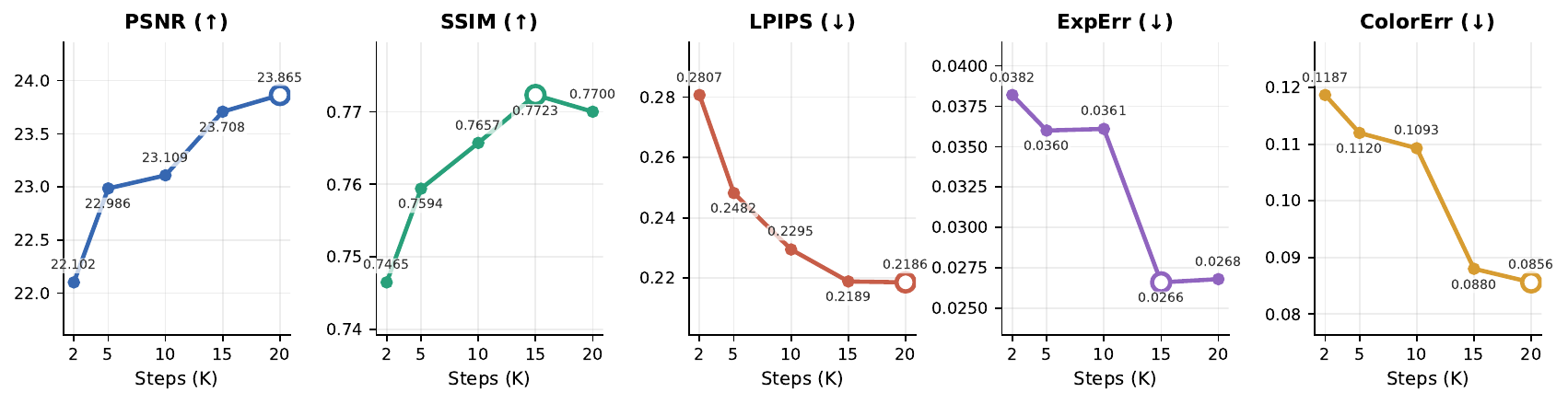}
	\vspace{-0.15cm}
\caption{
	Checkpoint sensitivity on \textbf{\textit{RELLISUR}} \(\times2\).
	Metrics become stable after 15K iterations, indicating that the final performance is not tied to a specific checkpoint.
}
	\label{fig:checkpoint_sensitivity}
	\vspace{-0.15cm}
\end{figure}

Qualitative error maps in Fig.~\ref{fig:ablation_roi_error} further show that removing visual tokens or attribute alignment causes weaker local texture recovery and larger reconstruction errors. Finally, Fig.~\ref{fig:checkpoint_sensitivity} shows that the metrics become stable after 15K iterations, indicating that the final performance is not tied to a specific checkpoint.

%
%
%
%
%
%

\definecolor{ablfull}{RGB}{235,248,236}
\definecolor{ablbest}{RGB}{255,238,200}
\definecolor{ablhead}{RGB}{245,247,250}
\newcommand{\ablbest}[1]{\cellcolor{ablbest}\textbf{#1}}
\newcommand{\yes}{\textcolor{green!55!black}{\cmark}}
\newcommand{\no}{\textcolor{red!65!black}{\xmark}}

\begin{table}[!t]
	\centering
	\setlength{\tabcolsep}{4.0pt}
	\renewcommand{\arraystretch}{0.90}
	\vspace{-0.1cm}
	\caption{
		Controlled ablation on \textbf{\textit{RELLISUR}} \(\times2\) under the same 5K-step training budget.
		\(\ell_d\), \(\mathbf{h}_v\), and \(\boldsymbol{\delta}\) denote degradation language, visual degradation tokens, and descriptor conditioning of the update-strength predictor, respectively.
	}
	\label{tab:component_ablation}
	\vspace{-0.2cm}
	\resizebox{0.95\linewidth}{!}{%
		\begin{tabular}{@{}lcccc|ccccc@{}}
			\toprule[1.1pt]
			\rowcolor{ablhead}
			\multirow{2}{*}{\textbf{\emph{Variant}}}
			& \multicolumn{4}{c|}{\textbf{\emph{Degradation-Control Signals}}}
			& \multicolumn{5}{c}{\textbf{\emph{Quantitative Metrics}}} \\
			\cmidrule(lr){2-5} \cmidrule(lr){6-10}
			\rowcolor{ablhead}
			&
			\(\ell_d\)
			& \(\mathbf{h}_{v}\)
			& \(\boldsymbol{\delta}\)
			& \textbf{Attr.}
			& \textbf{PSNR\(\uparrow\)}
			& \textbf{SSIM\(\uparrow\)}
			& \textbf{LPIPS\(\downarrow\)}
			& \textbf{ExpErr\(\downarrow\)}
			& \textbf{ColorErr\(\downarrow\)} \\
			\midrule
			
			Base model
			& \no & \no & \no & \no
			 & 21.358 & 0.7529 & 0.2796& 0.0531 & 0.1614 \\
			
			\midrule
			
			\rowcolor{ablfull}
			\textbf{Full DeLang-SR @5K}
			& \yes & \yes & \yes & \yes
			& \ablbest{22.986}
			& \ablbest{0.7594}
			& 0.2482
			& 0.0360
			& 0.1120 \\
			
			w/o degradation language
			& \no & \yes & \yes & \yes
			& 22.433
			& 0.7583
			& 0.2786
			& 0.0383
			& 0.1236 \\
			
			w/o visual degradation tokens
			& \yes & \no & \yes & \no
			& 22.398
			& 0.7500
			& 0.2589
			& 0.0423
			& 0.1270 \\
			
			w/o descriptor conditioning in \(\mathbf{M}_{\phi}\)
			& \yes & \yes & \no & \yes
			& 22.754
			& 0.7577
			& 0.2534
			& \ablbest{0.0335}
			& \ablbest{0.1023} \\
			
			w/o attribute alignment
			& \yes & \yes & \yes & \no
			& 22.516
			& 0.7539
			& \ablbest{0.2456}
			& 0.0425
			& 0.1274 \\
			
			\bottomrule[1.1pt]
		\end{tabular}
	}
	\vspace{-0.2cm}
\end{table}

\FloatBarrier

\section{Conclusion}

We presented \textbf{DeLang-SR}, a language-steered one-step flow-matching framework for super-resolution in the dark. By decoupling restoration direction and spatial update magnitude, DeLang-SR uses degradation language and visual tokens to steer the SD3 velocity field, while an update-strength field controls local Euler step sizes across heterogeneous dark regions. Experiments, ablations, and diagnostics demonstrate effective and interpretable dark super-resolution without oracle degradation annotations or iterative sampling. Nevertheless, DeLang-SR still inherits the computational footprint of a large pretrained SD3 prior and may hallucinate plausible details when scene structures are nearly invisible. Future work will explore lightweight flow priors, uncertainty-aware degradation estimation, and confidence-calibrated update-strength control.


\bibliographystyle{plainnat}
\bibliography{neurips_refs}

@inproceedings{saharia2022palette,
	title={Palette: Image-to-image diffusion models},
	author={Saharia, Chitwan and Chan, William and Chang, Huiwen and Lee, Chris and Ho, Jonathan and Salimans, Tim and Fleet, David and Norouzi, Mohammad},
	booktitle={ACM SIGGRAPH 2022 conference proceedings},
	pages={1--10},
	year={2022}
}

@inproceedings{wang2023zero,
	title={Zero-Shot Image Restoration Using Denoising Diffusion Null-Space Model},
	author={Wang, Yinhuai and Yu, Jiwen and Zhang, Jian},
	booktitle={ICLR},
	year={2023}
}

@inproceedings{fei2023generative,
	title={Generative diffusion prior for unified image restoration and enhancement},
	author={Fei, Ben and Lyu, Zhaoyang and Pan, Liang and Zhang, Junzhe and Yang, Weidong and Luo, Tianyue and Zhang, Bo and Dai, Bo},
	booktitle={CVPR},
	pages={9935--9946},
	year={2023}
}

@inproceedings{chen2018learning,
	title={Learning to see in the dark},
	author={Chen, Chen and Chen, Qifeng and Xu, Jia and Koltun, Vladlen},
	booktitle={CVPR},
	pages={3291--3300},
	year={2018}
}

@article{jiang2021enlightengan,
	title={Enlightengan: Deep light enhancement without paired supervision},
	author={Jiang, Yifan and Gong, Xinyu and Liu, Ding and Cheng, Yu and Fang, Chen and Shen, Xiaohui and Yang, Jianchao and Zhou, Pan and Wang, Zhangyang},
	journal={IEEE Transactions on Image Processing},
	volume={30},
	pages={2340--2349},
	year={2021}
}

@inproceedings{zamir2022restormer,
	title={Restormer: Efficient transformer for high-resolution image restoration},
	author={Zamir, Syed Waqas and Arora, Aditya and Khan, Salman and Hayat, Munawar and Khan, Fahad Shahbaz and Yang, Ming-Hsuan},
	booktitle={CVPR},
	pages={5728--5739},
	year={2022}
}

@inproceedings{wang2023promptir,
	title={{PromptIR}: Prompting for All-in-One Image Restoration},
	author={Potlapalli, Vaishnav and Zamir, Syed Waqas and Khan, Salman H and Shahbaz Khan, Fahad},
	booktitle={NeurIPS},
	pages={71275--71293},
	year={2023}
}

@inproceedings{lipman2022flow,
	title={Flow Matching for Generative Modeling},
	author={Lipman, Yaron and Chen, Ricky T. Q. and Ben-Hamu, Heli and Nickel, Maximilian and Le, Matt},
	booktitle={ICLR},
	year={2023}
}

@inproceedings{liu2022flow,
	title={Flow straight and fast: Learning to generate and transfer data with rectified flow},
	author={Liu, Xingchao and Gong, Chengyue and Liu, Qiang},
	booktitle={ICLR},
	year={2023}
}

@inproceedings{aakerberg2021rellisur,
	title={RELLISUR: A Real Low-Light Image Super-Resolution Dataset},
	author={Aakerberg, Andreas and Nasrollahi, Kamal and Moeslund, Thomas B.},
	booktitle={NeurIPS Datasets and Benchmarks Track},
	year={2021}
}

@inproceedings{hu2022lora,
	title={LoRA: Low-Rank Adaptation of Large Language Models},
	author={Hu, Edward J. and Shen, Yelong and Wallis, Phillip and Allen-Zhu, Zeyuan and Li, Yuanzhi and Wang, Shean and Wang, Lu and Chen, Weizhu},
	booktitle={ICLR},
	year={2022}
}

@inproceedings{Qin2025ReversingFF,
	title={Reversing flow for image restoration},
	author={Qin, Haina and Luo, Wenyang and Wang, Libin and Zheng, Dandan and Chen, Jingdong and Yang, Ming and Li, Bing and Hu, Weiming},
	booktitle={CVPR},
	pages={7545--7558},
	year={2025}
}

@inproceedings{luo2024daclip,
	title={Controlling Vision-Language Models for Multi-Task Image Restoration},
	author={Luo, Ziwei and Gustafsson, Fredrik K. and Zhao, Zheng and Sj{\"o}lund, Jens and Sch{\"o}n, Thomas B.},
	booktitle={ICLR},
	year={2024}
}

@article{simeoni2025dinov3,
	title={DINOv3},
	author={Sim{\'e}oni, Oriane and Vo, Huy V. and Seitzer, Maximilian and Baldassarre, Federico and Oquab, Maxime and Jose, Cijo and Khalidov, Vasil and Szafraniec, Marc and Yi, Seungeun and Ramamonjisoa, Micha{\"e}l and Massa, Francisco and Haziza, Daniel and Wehrstedt, Luca and Wang, Jianyuan and Darcet, Timoth{\'e}e and Moutakanni, Th{\'e}o and Sentana, Leonel and Roberts, Claire and Vedaldi, Andrea and Tolan, Jamie and Brandt, John and Couprie, Camille and Mairal, Julien and J{\'e}gou, Herv{\'e} and Labatut, Patrick and Bojanowski, Piotr},
	journal={arXiv preprint arXiv:2508.10104},
	year={2025}
}

@inproceedings{perez2018film,
	title={Film: Visual reasoning with a general conditioning layer},
	author={Perez, Ethan and Strub, Florian and De Vries, Harm and Dumoulin, Vincent and Courville, Aaron},
	booktitle={AAAI},
	volume={32},
	number={1},
	year={2018}
}

@inproceedings{park2025efficient,
	title={Efficient attention-sharing information distillation transformer for lightweight single image super-resolution},
	author={Park, Karam and Soh, Jae Woong and Cho, Nam Ik},
	booktitle={AAAI},
	volume={39},
	number={6},
	pages={6416--6424},
	year={2025}
}

@inproceedings{podell2023sdxl,
	title={SDXL: Improving Latent Diffusion Models for High-Resolution Image Synthesis},
	author={Podell, Dustin and English, Zion and Lacey, Kyle and Blattmann, Andreas and Dockhorn, Tim and M{\"u}ller, Jonas and Penna, Joe and Rombach, Robin},
	booktitle={ICLR},
	year={2024}
}

@article{wu2025omgsr,
	title={OMGSR: You Only Need One Mid-timestep Guidance for Real-World Image Super-Resolution},
	author={Wu, Zhiqiang and Sun, Zhaomang and Zhou, Tong and Fu, Bingtao and Cong, Ji and Dong, Yitong and Zhang, Huaqi and Tang, Xuan and Chen, Mingsong and Wei, Xian},
	journal={arXiv preprint arXiv:2508.08227},
	year={2025}
}

@inproceedings{esser2024sd3,
	title={Scaling rectified flow transformers for high-resolution image synthesis},
	author={Esser, Patrick and Kulal, Sumith and Blattmann, Andreas and Entezari, Rahim and M{\"u}ller, Jonas and Saini, Harry and Levi, Yam and Lorenz, Dominik and Sauer, Axel and Boesel, Frederic and others},
	booktitle={ICML},
	pages = {12606--12633},
	year={2024}
}

@article{poor_visibility_benchmark,
	title={Advancing image understanding in poor visibility environments: A collective benchmark study},
	author={Yang, Wenhan and Yuan, Ye and Ren, Wenqi and Liu, Jiaying and Scheirer, Walter J and Wang, Zhangyang and Zhang, Taiheng and Zhong, Qiaoyong and Xie, Di and Pu, Shiliang and others},
	journal={IEEE Transactions on Image Processing},
	volume={29},
	pages={5737--5752},
	year={2020} 
}

@article{wang2004image,
	title={Image Quality Assessment: From Error Visibility to Structural Similarity},
	author={Wang, Zhou and Bovik, Alan C. and Sheikh, Hamid R. and Simoncelli, Eero P.},
	journal={IEEE Transactions on Image Processing},
	volume={13},
	number={4},
	pages={600--612},
	year={2004}
}

@inproceedings{zhang2018perceptual,
	title={The Unreasonable Effectiveness of Deep Features as a Perceptual Metric},
	author={Zhang, Richard and Isola, Phillip and Efros, Alexei A. and Shechtman, Eli and Wang, Oliver},
	booktitle={CVPR},
	pages={586--595},
	year={2018}
}

@article{ding2020image,
	title={Image Quality Assessment: Unifying Structure and Texture Similarity},
	author={Ding, Keyan and Ma, Kede and Wang, Shiqi and Simoncelli, Eero P.},
	journal={IEEE Transactions on Pattern Analysis and Machine Intelligence},
	volume={44},
	number={5},
	pages={2567--2581},
	year={2022}
}

@article{mittal2012making,
	title={Making a “completely blind” image quality analyzer},
	author={Mittal, Anish and Soundararajan, Rajiv and Bovik, Alan C},
	journal={IEEE Signal Processing Letters},
	volume={20},
	number={3},
	pages={209--212},
	year={2012} 
}

@inproceedings{ke2021musiq,
	title={MUSIQ: Multi-Scale Image Quality Transformer},
	author={Ke, Junjie and Wang, Qifei and Wang, Yilin and Milanfar, Peyman and Yang, Feng},
	booktitle={ICCV},
	pages={5148--5157},
	year={2021}
}

@inproceedings{yang2022maniqa,
	title={MANIQA: Multi-Dimension Attention Network for No-Reference Image Quality Assessment},
	author={Yang, Sidi and Wu, Tianhe and Shi, Shuwei and Lao, Shanshan and Gong, Yuan and Cao, Mingdeng and Wang, Jiahao and Yang, Yujiu},
	booktitle={CVPR Workshops},
	pages={1190--1199},
	year={2022}
}

@inproceedings{wang2023clipiqa,
	title={Exploring CLIP for Assessing the Look and Feel of Images},
	author={Wang, Jianyi and Chan, Kelvin C. K. and Loy, Chen Change},
	booktitle={AAAI},
	volume={37},
	number={2},
	pages={2555--2563},
	year={2023}
}

@inproceedings{liu2025catanet,
	title={Catanet: Efficient content-aware token aggregation for lightweight image super-resolution},
	author={Liu, Xin and Liu, Jie and Tang, Jie and Wu, Gangshan},
	booktitle={CVPR},
	pages={17902--17912},
	year={2025}
}

@inproceedings{tang2026dod,
	title={Diffusion Once and Done: Degradation-Aware LoRA for All-in-One Image Restoration},
	author={Tang, Ni and Luo, Xiaotong and Cheng, Zihan and Zhou, Liangtai and Zhang, Dongxiao and Qu, Yanyun},
	booktitle={AAAI},
	volume={40},
	number={11},
	pages={9448--9456},
	year={2026}
}

@inproceedings{li2026seeing,
	title={Seeing through the rain: Resolving high-frequency conflicts in deraining and super-resolution via diffusion guidance},
	author={Li, Wenjie and Shi, Jinglei and Han, Jin and Guo, Heng and Ma, Zhanyu},
	booktitle={AAAI},
	volume={40},
	number={8},
	pages={6468--6476},
	year={2026}
}

@inproceedings{zhou2023srformer,
	title={Srformer: Permuted self-attention for single image super-resolution},
	author={Zhou, Yupeng and Li, Zhen and Guo, Chun-Le and Bai, Song and Cheng, Ming-Ming and Hou, Qibin},
	booktitle={CVPR},
	pages={12780--12791},
	year={2023}
}

@inproceedings{chen2023hat,
	title={Activating More Pixels in Image Super-Resolution Transformer},
	author={Chen, Xiangyu and Wang, Xintao and Zhou, Jiantao and Qiao, Yu and Dong, Chao},
	booktitle={CVPR},
	pages={22367--22377},
	year={2023}
}

@inproceedings{chen2024recursive,
	title={Recursive Generalization Transformer for Image Super-Resolution},
	author={Chen, Zheng and Zhang, Yulun and Gu, Jinjin and Kong, Linghe and Yang, Xiaokang},
	booktitle={ICLR},
	year={2024}
}

@inproceedings{long2025progressive,
	title={Progressive focused transformer for single image super-resolution},
	author={Long, Wei and Zhou, Xingyu and Zhang, Leheng and Gu, Shuhang},
	booktitle={CVPR},
	pages={2279--2288},
	year={2025}
}

@inproceedings{zhang2024hit,
	title={HiT-SR: Hierarchical Transformer for Efficient Image Super-Resolution},
	author={Zhang, Xiang and Zhang, Yulun and Yu, Fisher},
	booktitle={ECCV},
	pages={483--500},
	year={2024}
}

@inproceedings{liu2021retinex,
	title={Retinex-Inspired Unrolling with Cooperative Prior Architecture Search for Low-Light Image Enhancement},
	author={Liu, Risheng and Ma, Long and Zhang, Jiaao and Fan, Xin and Luo, Zhongxuan},
	booktitle={CVPR},
	pages={10561--10570},
	year={2021}
}

@inproceedings{wang2023ultra,
	title={Ultra-High-Definition Low-Light Image Enhancement: A Benchmark and Transformer-Based Method},
	author={Wang, Tao and Zhang, Kaihao and Shen, Tianrun and Luo, Wenhan and Stenger, Bjorn and Lu, Tong},
	booktitle={AAAI},
	volume={37},
	number={3},
	pages={2654--2662},
	year={2023}
}

@inproceedings{ma2022sci,
	title={Toward Fast, Flexible, and Robust Low-Light Image Enhancement},
	author={Ma, Long and Ma, Tengyu and Liu, Risheng and Fan, Xin and Luo, Zhongxuan},
	booktitle={CVPR},
	pages={5637--5646},
	year={2022}
}

@inproceedings{wang2021realesrgan,
	title={Real-ESRGAN: Training Real-World Blind Super-Resolution with Pure Synthetic Data},
	author={Wang, Xintao and Xie, Liangbin and Dong, Chao and Shan, Ying},
	booktitle={ICCV Workshops},
	pages={1905--1914},
	year={2021}
}

@inproceedings{liang2021swinir,
	title={SwinIR: Image Restoration Using Swin Transformer},
	author={Liang, Jingyun and Cao, Jiezhang and Sun, Guolei and Zhang, Kai and Van Gool, Luc and Timofte, Radu},
	booktitle={ICCV Workshops},
	pages={1833--1844},
	year={2021}
}

@inproceedings{yue2024unveiling,
	title={Unveiling Details in the Dark: Simultaneous Brightening and Zooming for Low-Light Image Enhancement},
	author={Yue, Ziyu and Gao, Jiaxin and Su, Zhixun},
	booktitle={AAAI},
	pages={6899--6907},
	year={2024}
}

@inproceedings{ho2020ddpm,
	title={Denoising Diffusion Probabilistic Models},
	author={Ho, Jonathan and Jain, Ajay and Abbeel, Pieter},
	booktitle={NeurIPS},
	volume={33},
	pages={6840--6851},
	year={2020}
}

@inproceedings{rombach2022ldm,
	title={High-Resolution Image Synthesis with Latent Diffusion Models},
	author={Rombach, Robin and Blattmann, Andreas and Lorenz, Dominik and Esser, Patrick and Ommer, Bjorn},
	booktitle={CVPR},
	pages={10684--10695},
	year={2022}
}

@article{wang2024stablesr,
	title={Exploiting diffusion prior for real-world image super-resolution},
	author={Wang, Jianyi and Yue, Zongsheng and Zhou, Shangchen and Chan, Kelvin CK and Loy, Chen Change},
	journal={International Journal of Computer Vision},
	volume={132},
	number={12},
	pages={5929--5949},
	year={2024}
}

@inproceedings{yue2023resshift,
	title={{ResShift}: Efficient Diffusion Model for Image Super-Resolution by Residual Shifting},
	author={Yue, Zongsheng and Wang, Jianyi and Loy, Chen Change},
	booktitle={NeurIPS},
	pages={13294--13307},
	year={2023}
}

@inproceedings{lin2024diffbir,
	title={Diffbir: Toward blind image restoration with generative diffusion prior},
	author={Lin, Xinqi and He, Jingwen and Chen, Ziyan and Lyu, Zhaoyang and Dai, Bo and Yu, Fanghua and Qiao, Yu and Ouyang, Wanli and Dong, Chao},
	booktitle={ECCV},
	pages={430--448},
	year={2024} 
}

@inproceedings{wang2024sinsr,
	title={SinSR: Diffusion-Based Image Super-Resolution in a Single Step},
	author={Wang, Yufei and Yang, Wenhan and Chen, Xinyuan and Wang, Yaohui and Guo, Lanqing and Chau, Lap-Pui and Liu, Ziwei and Qiao, Yu and Kot, Alex C. and Wen, Bihan},
	booktitle={CVPR},
	pages={25796--25805},
	year={2024}
}

@inproceedings{wu2024seesr,
	title={SeeSR: Towards Semantics-Aware Real-World Image Super-Resolution},
	author={Wu, Rongyuan and Yang, Tao and Sun, Lingchen and Zhang, Zhengqiang and Li, Shuai and Zhang, Lei},
	booktitle={CVPR},
	pages={25456--25467},
	year={2024}
}

@inproceedings{dong2025tsdsr,
	title={{TSD-SR}: One-Step Diffusion with Target Score Distillation for Real-World Image Super-Resolution},
	author={Dong, Linwei and Fan, Qingnan and Guo, Yihong and Wang, Zhonghao and Zhang, Qi and Chen, Jinwei and Luo, Yawei and Zou, Changqing},
	booktitle={CVPR},
	pages={23174--23184},
	year={2025}
}

@article{zhang2024s3diff,
	title={Degradation-Guided One-Step Image Super-Resolution with Diffusion Priors},
	author={Zhang, Aiping and Yue, Zongsheng and Pei, Renjing and Ren, Wenqi and Cao, Xiaochun},
	journal={arXiv preprint arXiv:2409.17058},
	year={2024}
}

@article{chen2025promptsr,
	title={Image super-resolution with text prompt diffusion},
	author={Chen, Zheng and Zhang, Yulun and Gu, Jinjin and Yuan, Xin and Kong, Linghe and Chen, Guihai and Yang, Xiaokang},
	journal={arXiv preprint arXiv:2311.14282},
	year={2023}
}

@inproceedings{luo2025defusion,
	title={Visual-instructed degradation diffusion for all-in-one image restoration},
	author={Luo, Wenyang and Qin, Haina and Chen, Zewen and Wang, Libin and Zheng, Dandan and Li, Yuming and Liu, Yufan and Li, Bing and Hu, Weiming},
	booktitle={CVPR},
	pages={12764--12777},
	year={2025}
}

@inproceedings{zheng2024sargd,
  title={Self-Adaptive Reality-Guided Diffusion for Artifact-Free Super-Resolution},
  author={Zheng, Qingping and Zheng, Ling and Guo, Yuanfan and Li, Ying and Xu, Songcen and Deng, Jiankang and Xu, Hang},
  booktitle={CVPR},
  pages={25806--25816},
  year={2024}
}

@inproceedings{yang2020fidelity,
  title={From Fidelity to Perceptual Quality: A Semi-Supervised Approach for Low-Light Image Enhancement},
  author={Yang, Wenhan and Wang, Shiqi and Fang, Yuming and Wang, Yue and Liu, Jiaying},
  booktitle={CVPR},
  pages={3063--3072},
  year={2020}
}

@article{hai2023r2rnet,
  title={R2RNet: Low-Light Image Enhancement via Real-Low to Real-Normal Network},
  author={Hai, Jiang and Xuan, Zhu and Han, Songchen and Yang, Ren and Hao, Yutong and Zou, Fengzhu and Lin, Fang},
  journal={Journal of Visual Communication and Image Representation},
  volume={90},
  pages={103712},
  year={2023}
}

@article{gao2024collaba,
  title={Collaborative Brightening and Amplification of Low-Light Imagery via Bi-Level Adversarial Learning},
  author={Gao, Jiaxin and Liu, Yaohua and Yue, Ziyu and Fan, Xin and Liu, Risheng},
  journal={Pattern Recognition},
  volume={154},
  pages={110558},
  year={2024}
}

@article{gao2025dualstream,
  title={A Dual-Stream-Modulated Learning Framework for Illuminating and Super-Resolving Ultra-Dark Images},
  author={Gao, Jiaxin and Yue, Ziyu and Liu, Yaohua and Xie, Sihan and Fan, Xin and Liu, Risheng},
  journal={IEEE Transactions on Neural Networks and Learning Systems},
  volume={36},
  number={4},
  pages={7500--7513},
  year={2025}
}

@inproceedings{gao2024trico,
  title={Enhancing Images with Coupled Low-Resolution and Ultra-Dark Degradations: A Tri-Level Learning Framework},
  author={Gao, Jiaxin and Liu, Yaohua},
  booktitle={ACM MM},
  pages={8642--8651},
  year={2024}
}

@article{yue2024lolisrflow,
  title={{LoLiSRFlow}: Joint Single Image Low-Light Enhancement and Super-Resolution via Cross-Scale Transformer-Based Conditional Flow},
  author={Yue, Ziyu and Gao, Jiaxin and Xie, Sihan and Liu, Yang and Su, Zhixun},
  journal={arXiv preprint arXiv:2402.18871},
  year={2024}
}

@article{gao2023hyperguidance,
  title={Learning Adaptive Hyper-Guidance via Proxy-Based Bilevel Optimization for Image Enhancement},
  author={Gao, Jiaxin and Liu, Xiaokun and Liu, Risheng and Fan, Xin},
  journal={The Visual Computer},
  volume={39},
  number={4},
  pages={1471--1484},
  year={2023}
}

@article{liu2024constraint,
  title={Learning with Constraint Learning: New Perspective, Solution Strategy and Various Applications},
  author={Liu, Risheng and Gao, Jiaxin and Liu, Xuan and Fan, Xin},
  journal={IEEE Transactions on Pattern Analysis and Machine Intelligence},
  volume={46},
  number={7},
  pages={5026--5043},
  year={2024}
}

@inproceedings{gao2026snoc,
  title={{SNOC}: Subtle Nested Objective Configuration for Joint Ultra-Low-Light Enhancement and Super-Resolution},
  author={Gao, Jiaxin and Liu, Yaohua and Cui, Danchen and Zhao, Zhihui},
  booktitle={ICMR},
  pages={2172--2181},
  year={2026}
}


\appendix

\section{Appendix Overview}
\label{app:overview}

This appendix provides additional technical details, implementation protocols, and supplementary experimental results for DeLang-SR. 
\appref{app:deg_language} details the construction of degradation conditions from measurable low-light statistics, including continuous descriptors, discrete degradation labels, prompt generation, and LQ-only inference-time estimation. 
\appref{app:architecture} describes the visual degradation token encoder, the spatial update-strength predictor, and the one-step flow-update convention used for direction--magnitude decoupling. 
\appref{app:training_inference} provides the training objectives and inference procedure. 
\appref{app:exp_details} summarizes dataset processing, evaluation metrics, baseline protocols, and runtime settings. 
Finally, \appref{app:additional_results} presents condition-source analysis, cross-dataset generalization, update-strength sensitivity and uniform-strength ablations, and qualitative comparisons.

\section{Details of Degradation Condition Construction}
\label{app:deg_language}
This section supplements Sec.~3.2 by providing concrete definitions of the degradation statistics, continuous degradation descriptor, discrete labels, prompt template, and LQ-only condition construction.

\subsection{Low-Light Degradation Statistics}
\label{app:deg_statistics}

For a paired training sample \((y,x)\), we compute a compact statistic vector
\(\mathbf{s}(y,x)=[\Delta_e,\Delta_c,\Delta_d,r_{\mathrm{dark}},n_{\mathrm{shadow}},u_{\mathrm{illum}},\Delta_{\mathrm{rgb}}]\).
Following the standard RGB luminance conversion, we use
\(Y(I)=0.299I_R+0.587I_G+0.114I_B\).
The detailed definitions are summarized in Table~\ref{tab:app_degradation_statistics}.

\begin{table}[!h]
	\centering
	\setlength{\tabcolsep}{4.2pt}
	\renewcommand{\arraystretch}{1.10}
	\caption{
		Low-light degradation statistics used for degradation-condition construction.
		\pairedtag{} and \inputtag{} denote paired-reference and input-only statistics, respectively.
	}
	\label{tab:app_degradation_statistics}
	\resizebox{0.98\linewidth}{!}{
		\begin{tabular}{@{}clll@{}}
			\toprule[1.1pt]
			\rowcolor{tabhead}
			\textbf{Type} & \textbf{Statistic} & \textbf{Definition} & \textbf{Interpretation} \\
			\midrule
			
			\rowcolor{pairrow}
			\pairedtag
			& \(\Delta_e\)
			& \(\mathrm{Mean}(Y(x))-\mathrm{Mean}(Y(y))\)
			& Exposure gap between the target and low-light input. \\
			
			\rowcolor{pairrow}
			\pairedtag
			& \(\Delta_c\)
			& \(\mathrm{Std}(Y(x))-\mathrm{Std}(Y(y))\)
			& Contrast degradation caused by insufficient illumination. \\
			
			\rowcolor{pairrow}
			\pairedtag
			& \(\Delta_d\)
			& \(\Delta_d=g(x)-g(y),~~g(I)=\mathrm{Mean}(|\nabla_xY(I)|)+\mathrm{Mean}(|\nabla_yY(I)|)\)
			& Gradient-based detail loss. \\
			
			\midrule
			
			\rowcolor{inputrow}
			\inputtag
			& \(r_{\mathrm{dark}}\)
			& \(\frac{1}{HW}\sum_{i,j}\mathbf{1}[Y(y)_{i,j}<\tau_d]\)
			& Ratio of dark pixels in the low-light input. \\
			
			\rowcolor{inputrow}
			\inputtag
			& \(n_{\mathrm{shadow}}\)
			& \(\mathrm{Mean}_{Y(y)<\tau_n}\big(|Y(y)-\mathcal{B}(Y(y))|\big)\)
			& Local residual fluctuation in dark regions. \\
			
			\rowcolor{inputrow}
			\inputtag
			& \(u_{\mathrm{illum}}\)
			& \(\mathrm{Std}\big(\mathcal{B}_{large}(Y(y))\big)\)
			& Large-scale illumination non-uniformity. \\
			
			\rowcolor{pairrow}
			\pairedtag
			& \(\Delta_{\mathrm{rgb}}\)
			& \(\mathrm{Mean}_{rgb}(x)-\mathrm{Mean}_{rgb}(y)\)
			& RGB color shift between the target and low-light input. \\
			
			\bottomrule[1.1pt]
		\end{tabular}
	}
\end{table}

\subsection{Continuous Descriptor and Discrete Labels}
\label{app:descriptor_labels}

The statistics are converted into two complementary representations in Table~\ref{tab:app_descriptor_labels}: a continuous descriptor \(\boldsymbol{\delta}\) for update-strength prediction and discrete labels \(\boldsymbol{\kappa}\) for prompt generation and attribute alignment.
The continuous descriptor is
\[
\boldsymbol{\delta}
=
[
\delta_{\mathrm{exp}},
\delta_{\mathrm{dark}},
\delta_{\mathrm{noise}},
\delta_{\mathrm{contrast}},
\delta_{\mathrm{detail}},
\delta_{\mathrm{illum}},
\delta_{\mathrm{scale}}
]
\in\mathbb{R}^{7}.
\]
In practice, it is instantiated as
\[
\boldsymbol{\delta}
=
[
\bar{\Delta}_e,
r_{\mathrm{dark}},
\bar{n}_{\mathrm{shadow}},
\bar{\Delta}_c,
\bar{\Delta}_d,
\mathbf{1}_{\mathrm{nonuniform}},
r_{\mathrm{scale}}
],
\]
where bars denote percentile-normalized statistics:
\[
\bar{s}=\mathrm{clip}\left(
\frac{s-P_{low}(s)}
{P_{high}(s)-P_{low}(s)+\epsilon},
0,1
\right).
\]
The discrete labels are
\[
\boldsymbol{\kappa}
=
\{
\kappa_{\mathrm{exp}},
\kappa_{\mathrm{noise}},
\kappa_{\mathrm{contrast}},
\kappa_{\mathrm{detail}},
\kappa_{\mathrm{illum}},
\kappa_{\mathrm{color}}
\},
\]
where each label is obtained by quantizing the corresponding statistic using fixed thresholds or dataset-level percentiles.
Color cast is represented as the discrete label \(\kappa_{\mathrm{color}}\) rather than a continuous descriptor entry.

\begin{table}[!t]
	\centering
	\setlength{\tabcolsep}{4.4pt}
	\renewcommand{\arraystretch}{1.10}
	\caption{
		Continuous descriptor entries and discrete degradation labels.
		The descriptor \(\boldsymbol{\delta}\) guides update-strength prediction, while \(\boldsymbol{\kappa}\) fills degradation-language prompt slots and supervises attribute alignment.
	}
	\label{tab:app_descriptor_labels}
	\resizebox{0.98\linewidth}{!}{
		\begin{tabular}{@{}llll@{}}
			\toprule[1.1pt]
			\rowcolor{tabhead}
			\textbf{Degradation Factor}
			& \textbf{Continuous Descriptor}
			& \textbf{Discrete Label / Slot}
			& \textbf{Role} \\
			\midrule
			
			\rowcolor{pairrow}
			Underexposure
			& \(\delta_{\mathrm{exp}}\leftarrow\bar{\Delta}_e\)
			& \(\kappa_{\mathrm{exp}}\)
			& Exposure intent and visibility recovery. \\
			
			\rowcolor{inputrow}
			Dark regions
			& \(\delta_{\mathrm{dark}}\leftarrow r_{\mathrm{dark}}\)
			& --
			& Local darkness-aware magnitude control. \\
			
			\rowcolor{inputrow}
			Shadow noise
			& \(\delta_{\mathrm{noise}}\leftarrow\bar{n}_{\mathrm{shadow}}\)
			& \(\kappa_{\mathrm{noise}}\)
			& Noise-aware restoration intent. \\
			
			\rowcolor{pairrow}
			Contrast loss
			& \(\delta_{\mathrm{contrast}}\leftarrow\bar{\Delta}_c\)
			& \(\kappa_{\mathrm{contrast}}\)
			& Contrast restoration intent. \\
			
			\rowcolor{pairrow}
			Detail degradation
			& \(\delta_{\mathrm{detail}}\leftarrow\bar{\Delta}_d\)
			& \(\kappa_{\mathrm{detail}}\)
			& Detail recovery intent. \\
			
			\rowcolor{inputrow}
			Illumination non-uniformity
			& \(\delta_{\mathrm{illum}}\leftarrow\mathbf{1}_{\mathrm{nonuniform}}\)
			& \(\kappa_{\mathrm{illum}}\)
			& Non-uniform illumination control. \\
			
			\rowcolor{slotrow}
			Color cast
			& --
			& \(\kappa_{\mathrm{color}}\)
			& Color-cast prompt slot. \\
			
			\rowcolor{slotrow}
			Scale factor
			& \(\delta_{\mathrm{scale}}\leftarrow r_{\mathrm{scale}}\)
			& \prompttag{} prompt slot
			& Target SR scale instruction. \\
			
			\bottomrule[1.1pt]
		\end{tabular}
	}
\end{table}

\subsection{Prompt Generation and LQ-Only Inference}
\label{app:prompt_lq_only}

 The degradation prompt \(\ell_d\) is generated from the discrete labels using a deterministic template, i.e., \(\ell_d=\mathcal{T}(\boldsymbol{\kappa})\). 
Specifically, the discrete labels \(\boldsymbol{\kappa}\) fill the degradation-specific slots in the prompt template, as illustrated below.

\begin{center}
	\fcolorbox{black!25}{black!3}{
		\begin{minipage}{0.92\linewidth}
			\small
			\textbf{Degradation template.} 
			Restore a real low-light image with \texttt{[exposure]} underexposure, \texttt{[noise]} shadow noise, \texttt{[contrast]} contrast loss, \texttt{[detail]} detail degradation, \texttt{[illumination]} illumination, and a \texttt{[color]} color cast. 
			Perform \texttt{[scale]}\(\times\) super-resolution while preserving scene structure and natural colors.
			
			\vspace{0.4em}
			\textbf{Fixed task instruction.}
			Enhance dark regions, suppress shadow artifacts, recover fine details, and avoid over-exposure or hallucinated content.
		\end{minipage}
	}
\end{center}

For example, a \(4\times\) input with severe degradation may generate:

\begin{center}
	\fcolorbox{blue!35}{blue!3}{
		\begin{minipage}{0.92\linewidth}
			\small
			\textbf{Instantiated prompt.} Restore a real low-light image with severe underexposure, high shadow noise, strong contrast loss, severe detail degradation, non-uniform illumination, and a warm color cast. 
			Perform \(4\times\) super-resolution while preserving scene structure and natural colors. 
			Enhance dark regions, suppress shadow artifacts, recover fine details, and avoid over-exposure or hallucinated content.
		\end{minipage}
	}
\end{center}

The prompt is encoded by the frozen SD3 text encoder~\cite{esser2024sd3}: \(\mathbf{h}_{\ell}=E_{\mathrm{text}}(\ell_d)\). 
For efficiency, prompt embeddings can be cached offline during training.

\subsection{LQ-Only Condition Construction at Inference}
\label{app:lq_only}

At inference time, the high-quality reference \(x\) is unavailable. 
We therefore use an LQ-only construction function \((\boldsymbol{\delta},\ell_d)=\mathcal{V}_{test}(y)\). 
Exposure severity is estimated from the mean luminance and dark-region ratio of \(y\); contrast and detail degradation are estimated from luminance statistics and gradient responses; shadow noise and illumination non-uniformity are computed from the input-only statistics defined in Table~\ref{tab:app_degradation_statistics}; and color cast is estimated from the mean RGB distribution of \(y\). 
For statistics that require the paired target during training, the LQ-only variant uses the corresponding input-only proxy estimated from \(y\). 
This protocol requires no high-quality reference image or oracle degradation label during deployment. 

\section{Architecture and Flow-Update Details}
\label{app:architecture}

This section supplements Sec.~3.3 and Sec.~3.4 by providing implementation details of the visual degradation token encoder, the spatial update-strength predictor, and the one-step flow-update convention. 
DeLang-SR is built upon a pretrained SD3 prior~\cite{esser2024sd3}, whose rectified-flow formulation follows flow matching and rectified flow~\cite{lipman2022flow,liu2022flow}. 
We adapt the pretrained prior with lightweight LoRA modules~\cite{hu2022lora}, while the newly introduced degradation-control modules are trained from scratch.

\subsection{Visual Degradation Token Encoder}
\label{app:vdt}

The visual degradation token encoder \(G_\psi\) produces image-specific tokens that complement the global degradation-language intent. 
While the degradation prompt \(\ell_d\) specifies what degradations should be corrected, the visual tokens provide instance-level evidence about where degradations appear in the current input. 
Given the low-light input \(y\), we compute the luminance map \(Y(y)\), darkness map \(d_y=1-Y(y)\), and local contrast map \(\ell_y=|Y(y)-\mathrm{AvgPool}(Y(y))|\). 
These cues are concatenated with the input image to form a local evidence tensor \(\tilde{e}_y=[y,Y(y),d_y,\ell_y]\). 
A lightweight encoder \(q_\psi\) extracts degradation-aware features \(\mathbf{F}_{\mathrm{tok}}=q_\psi(\tilde{e}_y)\). 
The pooled feature is concatenated with global statistics \(\mathbf{r}(y)\), yielding \(\mathbf{u}_{\mathrm{tok}}=[\mathrm{Pool}(\mathbf{F}_{\mathrm{tok}});\mathbf{r}(y)]\), and is then projected into \(M\) visual degradation tokens:
\begin{equation}
	\mathbf{h}_{v}
	=
	G_\psi(y)
	=
	\mathrm{reshape}
	\left(
	\mathrm{MLP}(\mathbf{u}_{\mathrm{tok}})
	\right)
	\in\mathbb{R}^{M\times d_t}.
	\label{eq:app_token_encoder}
\end{equation}
Here \(d_t\) denotes the SD3 text-token dimension. 
We use \(M=4\) tokens to keep the additional sequence length negligible. 
The resulting \(\mathbf{h}_v\) is concatenated with the degradation-language embedding \(\mathbf{h}_{\ell}\) to form \(\mathbf{H}_d=[\mathbf{h}_{\ell};\mathbf{h}_{v}]\), which steers the SD3 velocity direction.

\subsection{Spatial Update-Strength Predictor}
\label{app:sigma_modulator}

The spatial update-strength predictor estimates \(\mathbf{M}_{\phi}(y,\boldsymbol{\delta})\), which serves as a locally varying Euler step-size field in the one-step flow update. 
Different from the visual token encoder that provides conditioning tokens for direction steering, this predictor controls how strongly the predicted velocity is applied at each latent location. 
In Eq.~\eqref{eq:sigma_prediction}, \(E_{\phi}(y)\) denotes the local visual feature extractor and \(P_{\phi}(\boldsymbol{\delta})\) denotes the descriptor-conditioned modulation branch. 
Concretely, we implement \(E_{\phi}(y)\) by constructing \(e_y=[y,d_y,\ell_y]\) and extracting \(\mathbf{F}_y=f_{\phi}(e_y)\). 
The descriptor branch \(P_{\phi}(\boldsymbol{\delta})\) predicts FiLM parameters~\cite{perez2018film}, i.e., \((\gamma_{\delta},\beta_{\delta})=P_{\phi}(\boldsymbol{\delta})\), which modulate the local visual feature as
\begin{equation}
	\widetilde{\mathbf{F}}_y
	=
	\mathbf{F}_y\odot(1+\gamma_{\delta})+\beta_{\delta}.
	\label{eq:app_film_modulation}
\end{equation}
The residual form \(1+\gamma_{\delta}\) keeps the modulation close to identity at initialization and stabilizes training. 
The spatial update-strength field is then predicted by
\begin{equation}
	\mathbf{M}_{\phi}(y,\boldsymbol{\delta})
	=
	\sigma_{\min}
	+
	(\sigma_{\max}-\sigma_{\min})
	\cdot
	\mathrm{Sigmoid}
	\left(
	g_\phi(\widetilde{\mathbf{F}}_y)
	\right).
	\label{eq:app_sigma}
\end{equation}
The field is resized to the SD3 latent resolution and broadcast along latent channels before multiplying the predicted velocity field. 
This design allows stronger correction in heavily degraded regions and more conservative updates in relatively reliable regions.

\subsection{One-Step Flow Update and Operating Timestep}
\label{app:flow_update}

Under the SD3 flow-matching formulation~\cite{esser2024sd3,lipman2022flow,liu2022flow}, the transformer predicts a velocity field conditioned on a latent state, timestep, and context. 
In DeLang-SR, we use the degraded latent \(z_y\) as the starting state and perform a single direction-steered and magnitude-controlled Euler update:
\begin{equation}
	\hat{z}
	=
	z_y
	-
	\mathbf{M}_{\phi}(y,\boldsymbol{\delta})
	\odot
	\mathbf{v}_{\theta}(z_y,t_m,\mathbf{H}_{d}).
	\label{eq:app_one_step_update}
\end{equation}
The sign follows the velocity convention used by the SD3 scheduler. 
In this update, \(\mathbf{H}_{d}\) determines the restoration direction of the predicted velocity field, while \(\mathbf{M}_{\phi}(y,\boldsymbol{\delta})\) controls the local update magnitude.

The fixed operating timestep \(t_m\) provides a stable point on the pretrained SD3 flow trajectory. 
A very small timestep may lead to weak correction, while a very large timestep may over-activate the generative prior and introduce hallucinated details. 
In our experiments, \(t_m\) is fixed for both training and inference, so image- and region-dependent restoration strength is controlled by \(\mathbf{M}_{\phi}(y,\boldsymbol{\delta})\) rather than by changing the operating timestep.

\subsection{Frozen and Trainable Components}
\label{app:frozen_trainable}

DeLang-SR adapts a pretrained SD3 prior~\cite{esser2024sd3} in a parameter-efficient manner. 
The SD3 text encoder is frozen to preserve its language-conditioning space, and the VAE decoder is frozen to preserve the pretrained decoding space. 
We insert lightweight LoRA adapters~\cite{hu2022lora} into the VAE encoder and SD3 flow transformer to adapt the latent representation and velocity prediction to dark super-resolution. 
The visual degradation token encoder \(G_\psi\), spatial update-strength predictor \(\mathbf{M}_{\phi}\), and attribute prediction head are trained from scratch. 
This design preserves the general generative prior of SD3 while enabling task-specific direction steering and magnitude control for super-resolution in the dark.

\section{Training Objective and Inference Procedure}
\label{app:training_inference}

This section supplements Sec.~3.4 by describing the loss terms used for training.
\subsection{Training Objective Details}
\label{app:objective_details}

The training objective supervises DeLang-SR from three complementary aspects: 
(i) latent flow alignment, which makes the adapted degraded latent compatible with the spatially modulated flow state; 
(ii) image reconstruction, which constrains the restored image in pixel, frequency, and perceptual spaces; 
and (iii) degradation-control regularization, which calibrates the spatial update-strength field and visual degradation tokens.
For perceptual supervision, we compute a DISTS-style distance~\cite{ding2020image} using DINOv3 visual features~\cite{simeoni2025dinov3}.

For each paired sample \((y,x)\), we construct a spatially degradation-dependent latent state:
\begin{equation}
	z_{\mathbf{M}}
	=
	(1-\mathbf{M}_{\phi}(y,\boldsymbol{\delta}))\odot z_x
	+
	\mathbf{M}_{\phi}(y,\boldsymbol{\delta})\odot \epsilon,
	\qquad
	\epsilon\sim\mathcal{N}(0,I).
	\label{eq:app_zsigma}
\end{equation}
This state provides a spatial flow-alignment target, where \(\mathbf{M}_{\phi}(y,\boldsymbol{\delta})\) determines the local interpolation strength between the high-quality latent \(z_x\) and random noise \(\epsilon\).

The loss terms are summarized in Table~\ref{tab:app_loss_terms}. 
Here, \(\mathrm{Low}(\cdot)\) denotes Gaussian low-pass filtering and \(\mathrm{High}(I)=I-\mathrm{Low}(I)\). 
The update-strength prior is defined as 
\(\mathbf{M}_{prior}(y)=\sigma_{\min}+(\sigma_{\max}-\sigma_{\min})d_y\), 
and \(K=6\) corresponds to the six discrete degradation attributes.

\begin{table}[!h]
	\centering
	\setlength{\tabcolsep}{4.2pt}
	\renewcommand{\arraystretch}{1.08}
	\caption{
		Summary of training objectives used in DeLang-SR.
		The losses supervise latent flow alignment, image reconstruction, and degradation-control regularization.
	}
	\label{tab:app_loss_terms}
	\resizebox{0.98\linewidth}{!}{
		\begin{tabular}{@{}llll@{}}
			\toprule[1.1pt]
			\rowcolor{losshead}
			\textbf{Group} & \textbf{Term} & \textbf{Definition} & \textbf{Role} \\
			\midrule
			
			\rowcolor{losslatent}
			Latent alignment
			& \(\mathcal{L}_{lat}\)
			& \(\|z_y-z_{\mathbf{M}}\|_2^2\)
			& Aligns the degraded latent with the spatially modulated flow state. \\
			
			\midrule
			
			\rowcolor{lossrecon}
			Pixel reconstruction
			& \(\mathcal{L}_{pix}\)
			& \(\|\hat{x}-x\|_1\)
			& Supervises pixel-level fidelity. \\
			
			\rowcolor{lossrecon}
			Frequency reconstruction
			& \(\mathcal{L}_{freq}\)
			& \(\|\mathrm{Low}(\hat{x})-\mathrm{Low}(x)\|_1
			+
			\|\mathrm{High}(\hat{x})-\mathrm{High}(x)\|_1\)
			& Constrains illumination/color and high-frequency details. \\
			
			\rowcolor{lossrecon}
			Perceptual consistency
			& \(\mathcal{L}_{perc}\)
			& \(\mathrm{DISTS}_{DINOv3}(\hat{x},x)\)
			& Uses DISTS-style supervision with DINOv3 features~\cite{ding2020image,simeoni2025dinov3}. \\
			
			\midrule
			
			\rowcolor{losscontrol}
			Update-strength prior
			& \(\mathcal{L}_{M}\)
			& \(\|\mathbf{M}_{\phi}(y,\boldsymbol{\delta})-\mathbf{M}_{prior}(y)\|_1\)
			& Calibrates the spatial update-strength field. \\
			
			\rowcolor{losscontrol}
			Attribute alignment
			& \(\mathcal{L}_{attr}\)
			& \(\sum_{k=1}^{K}\mathrm{CE}(\hat{\kappa}^{k},\kappa^{k})\)
			& Aligns visual tokens with discrete degradation labels. \\
			
			\bottomrule[1.1pt]
		\end{tabular}
	}
\end{table}
%

\subsection{Training and Inference Algorithms}
\label{app:train_infer_alg}

Algorithm~\ref{alg:training} summarizes the training procedure of DeLang-SR. 
During training, paired samples are used to construct degradation conditions and supervise reconstruction, while the SD3 text encoder and VAE decoder remain frozen. 
Algorithm~\ref{alg:inference} summarizes the inference procedure, where only the low-light input \(y\) is required.

\begin{algorithm}[!h]
	\caption{Training procedure of DeLang-SR}
	\label{alg:training}
	\begin{algorithmic}[1]
		\REQUIRE Paired training data \(\{(y,x)\}\), fixed timestep \(t_m\), pretrained SD3 prior.
		\ENSURE Trainable parameters of LoRA adapters, visual token encoder \(G_{\psi}\), update-strength predictor \(\mathbf{M}_{\phi}\), and attribute head \(H\).
		
		\STATE \textbf{Initialization.}
		Freeze the SD3 text encoder \(E_{\mathrm{text}}\) and VAE decoder \(\mathcal{D}_{vae}\); insert LoRA adapters into the VAE encoder and SD3 flow transformer.
		
		\FOR{each mini-batch \((y,x)\)}
		
		\STATE \textbf{Degradation-condition construction.}
		Construct degradation conditions
		\[
		(\boldsymbol{\delta},\boldsymbol{\kappa},\ell_d)=\mathcal{V}(y,x).
		\]
		
		\STATE \textbf{Direction steering.}
		Encode degradation language and visual evidence:
		\[
		\mathbf{h}_{\ell}=E_{\mathrm{text}}(\ell_d), 
		\qquad
		\mathbf{h}_{v}=G_{\psi}(y), 
		\qquad
		\mathbf{H}_{d}=[\mathbf{h}_{\ell};\mathbf{h}_{v}].
		\]
		
		\STATE \textbf{Magnitude control.}
		Encode latents and predict the update-strength field:
		\[
		z_x=\mathcal{E}_{fix}(x), 
		\qquad
		z_y=\mathcal{E}_{adapt}(y), 
		\qquad
		\mathbf{M}=\mathbf{M}_{\phi}(y,\boldsymbol{\delta}).
		\]
		
		\STATE \textbf{One-step flow restoration.}
		Predict the SD3 velocity field and perform the direction-steered, magnitude-controlled update:
		\[
		\mathbf{v}=\mathbf{v}_{\theta}(z_y,t_m,\mathbf{H}_{d}),
		\qquad
		\hat{z}=z_y-\mathbf{M}\odot\mathbf{v},
		\qquad
		\hat{x}=\mathcal{D}_{vae}(\hat{z}).
		\]
		
		\STATE \textbf{Objective computation.}
		Construct the spatial latent target
		\[
		z_{\mathbf{M}}=(1-\mathbf{M})\odot z_x+\mathbf{M}\odot\epsilon,
		\qquad
		\epsilon\sim\mathcal{N}(0,I),
		\]
		and compute \(\mathcal{L}_{lat}\), \(\mathcal{L}_{pix}\), \(\mathcal{L}_{freq}\), \(\mathcal{L}_{perc}\), \(\mathcal{L}_{M}\), and \(\mathcal{L}_{attr}\).
		
		\STATE \textbf{Parameter update.}
		Update all trainable parameters by minimizing the total objective in Eq.~\eqref{eq:total_objective}.
		
		\ENDFOR
	\end{algorithmic}
\end{algorithm}

\begin{algorithm}[!h]
	\caption{Inference procedure of DeLang-SR}
	\label{alg:inference}
	\begin{algorithmic}[1]
		\REQUIRE Low-light low-resolution input \(y\), fixed timestep \(t_m\).
		\ENSURE Restored high-resolution image \(\hat{x}\).
		
		\STATE \textbf{LQ-only condition construction.}
		Estimate degradation conditions from the input only:
		\[
		(\boldsymbol{\delta},\ell_d)=\mathcal{V}_{test}(y).
		\]
		
		\STATE \textbf{Direction steering.}
		Encode degradation language and visual evidence:
		\[
		\mathbf{h}_{\ell}=E_{\mathrm{text}}(\ell_d),
		\qquad
		\mathbf{h}_{v}=G_{\psi}(y),
		\qquad
		\mathbf{H}_{d}=[\mathbf{h}_{\ell};\mathbf{h}_{v}].
		\]
		
		\STATE \textbf{Magnitude control.}
		Encode the input latent and predict the spatial update-strength field:
		\[
		z_y=\mathcal{E}_{adapt}(y),
		\qquad
		\mathbf{M}=\mathbf{M}_{\phi}(y,\boldsymbol{\delta}).
		\]
		
		\STATE \textbf{One-step flow restoration.}
		Predict the SD3 velocity field and perform the final update:
		\[
		\mathbf{v}=\mathbf{v}_{\theta}(z_y,t_m,\mathbf{H}_{d}),
		\qquad
		\hat{z}=z_y-\mathbf{M}\odot\mathbf{v},
		\qquad
		\hat{x}=\mathcal{D}_{vae}(\hat{z}).
		\]
		
		\RETURN \(\hat{x}\).
	\end{algorithmic}
\end{algorithm}

\section{Implementation Details}
\label{app:implementation}

DeLang-SR is implemented based on a pretrained Stable Diffusion 3 (SD3) model~\cite{esser2024sd3}. The SD3 text encoder and VAE decoder are frozen. The VAE encoder and SD3 flow transformer are adapted with LoRA modules~\cite{hu2022lora}. The visual degradation token encoder, spatial update-strength predictor, and attribute prediction head are randomly initialized and optimized jointly with the LoRA adapters. We use \(M=4\) visual degradation tokens and a 7-dimensional continuous degradation descriptor. The spatial update-strength field is bounded by \(\sigma_{\min}=0.15\) and \(\sigma_{\max}=0.95\), resized to the latent resolution, and broadcast along latent channels before multiplying the predicted flow. During training, degradation prompts are generated from metadata and fixed task instructions, and their text embeddings can be cached for efficiency.

The loss weights are \(\lambda_{lat}=5\), \(\lambda_{perc}=5\), \(\lambda_{pix}=0.7\), \(\lambda_M=0.2\), and \(\lambda_{attr}=0.1\); the low- and high-frequency terms are each weighted by $0.5$. The LoRA ranks for the VAE encoder and flow transformer are 96 and 128, respectively. For both adapters, the scaling factor \(\alpha\) is set to the rank, the weights are initialized with a Gaussian distribution, and no LoRA dropout is applied. The VAE adapters target convolutional and attention projections, while the transformer adapters target the input embedding, attention projections, and feed-forward layers.

For controlled ablations, we train all variants on RELLISUR \(\times2\) under the same 5K-step budget. The full model used for main comparison and diagnostic visualization is selected according to validation performance. Runtime and memory are measured during inference, excluding model loading and metric computation.

\section{Experimental Details}
\label{app:exp_details}

\subsection{Dataset Processing}
\label{app:dataset_processing}

For paired evaluation, we use RELLISUR~\cite{aakerberg2021rellisur}, which contains real low-light low-resolution images paired with normal-light high-resolution references. Following the standard split, we use 3,610 paired samples for training and 425 paired samples for testing under both \(\times 2\) and \(\times 4\) settings. The low-light image is denoted as \(y\), and the paired normal-light image is denoted as \(x\). For real-scene generalization, we additionally evaluate on 100 images selected from DarkFace~\cite{poor_visibility_benchmark}. Since DarkFace does not provide paired high-quality references, it is used only for no-reference perceptual evaluation and visual comparison.

We additionally evaluate cross-dataset transfer on 100 LOL-v2-Real test pairs and 50 LSRW test pairs. Because both datasets were introduced for paired low-light enhancement rather than super-resolution, we construct a \(2\times\) dark-SR protocol by bicubically downsampling each low-light image and retaining the paired normal-light image as the high-resolution target. All methods use models trained on RELLISUR and are evaluated without fine-tuning on LOL-v2-Real or LSRW.

All methods are evaluated at the same output resolution. For paired RELLISUR evaluation, full-reference metrics are computed between the restored image \(\hat{x}\) and the paired target \(x\). For DarkFace, no-reference metrics are computed directly on the restored images. Unless otherwise specified, no high-quality reference image, oracle degradation label, or manual prompt is used during inference.

\subsection{Evaluation Metrics}
\label{app:metric_details}

For paired evaluation, we report PSNR, SSIM~\cite{wang2004image}, LPIPS~\cite{zhang2018perceptual}, and DISTS~\cite{ding2020image}. PSNR and SSIM measure pixel-level fidelity and structural similarity, while LPIPS and DISTS evaluate perceptual and semantic consistency. To better reflect low-light restoration quality, we further report exposure error and color consistency error. Given the restored image \(\hat{x}\) and target image \(x\), exposure error is defined as
\begin{equation}
	\mathrm{ExpErr}
	=
	\left|
	\mathrm{Mean}(Y(\hat{x}))
	-
	\mathrm{Mean}(Y(x))
	\right|,
\end{equation}
where \(Y(\cdot)\) denotes the luminance channel. The color consistency error is defined as
\begin{equation}
	\mathrm{ColorErr}
	=
	\left\|
	\mathrm{Mean}_{rgb}(\hat{x})
	-
	\mathrm{Mean}_{rgb}(x)
	\right\|_1.
\end{equation}
These two metrics are used as task-specific diagnostic measures for illumination naturalness and chromatic fidelity.

For no-reference perceptual evaluation, we report NIQE~\cite{mittal2012making}, MUSIQ~\cite{ke2021musiq}, MANIQA~\cite{yang2022maniqa}, and CLIP-IQA~\cite{wang2023clipiqa}. NIQE is lower better, while MUSIQ, MANIQA, and CLIP-IQA are higher better.

\subsection{Baseline Evaluation Protocol}
\label{app:baseline_protocol}

We evaluate three groups of baselines. For general SR/restoration backbones, low-light inputs are directly fed into the corresponding models and evaluated at the target output resolution. For cascaded LLE+SR pipelines, we consider both enhancement-before-SR and SR-before-enhancement orders when applicable. Low-light enhancement models are trained or applied on the \(1\times\) setting, while SR models are trained or evaluated under the corresponding \(\times 2\) or \(\times 4\) scale. For diffusion-guided or degradation-guided methods, we follow their official inference configurations when available and resize the outputs to match the evaluation resolution. All reported metrics are computed using the same test images and evaluation scripts.

\subsection{Training and Runtime Settings}
\label{app:training_runtime}

DeLang-SR is implemented in PyTorch. The model is initialized from a pretrained SD3 prior~\cite{esser2024sd3}. The SD3 text encoder and VAE decoder are frozen. The VAE encoder and SD3 flow transformer are adapted with LoRA modules, while \(G_\psi\), \(\mathbf{M}_{\phi}\), and the attribute prediction head are trained from scratch. We use \(M=4\) visual degradation tokens and a 7-dimensional degradation descriptor \(\boldsymbol{\delta}\). During inference, text embeddings can be cached once the degradation prompt \(\ell_d\) is generated. DeLang-SR performs restoration with a single SD3 flow update, without iterative diffusion sampling. 

We train the \(\times2\) and \(\times4\) models at resolutions of \(1024\times1024\) and \(864\times864\), respectively, using AdamW with a learning rate of \(5\times10^{-5}\), \(\beta_1=0.9\), \(\beta_2=0.999\), weight decay \(10^{-4}\), and gradient clipping with a maximum norm of 1.0. 
The per-device batch size is 1 and gradient accumulation is set to 2, giving an effective batch size of 2 with one training GPU. 
Both the \(\times2\) and \(\times4\) models are trained for 20K optimization steps with mixed-precision training, and checkpoints are saved every 1K steps. 
We use LoRA ranks 96 and 128 for the SD3 VAE encoder and transformer, respectively, while the pretrained SD3 components are otherwise kept frozen.
All methods are evaluated at the same output resolution. 
Runtime and peak memory are measured with batch size 1 after warm-up, excluding model loading, prompt-cache construction, image I/O, and metric computation. 
The checkpoint used for the main comparison is selected according to validation performance.

\section{Additional Experimental Results}
\label{app:additional_results}

\subsection{Cross-Dataset Evaluation without Fine-Tuning}
\label{app:cross_dataset}

LOL-v2-Real~\cite{yang2020fidelity} and LSRW~\cite{hai2023r2rnet} do not provide native SR protocols. We therefore follow the construction described in \appref{app:dataset_processing} and evaluate the RELLISUR-trained models without further optimization. Tables~\ref{tab:cross_dataset_fr} and~\ref{tab:cross_dataset_nr} report full-reference and no-reference results, respectively.

\begin{table}[!h]
	\centering
	\setlength{\tabcolsep}{2.7pt}
	\renewcommand{\arraystretch}{0.96}
	\caption{Full-reference cross-dataset evaluation under the constructed \(\times2\) dark-SR protocol. All models are trained on RELLISUR and evaluated without fine-tuning.}
	\label{tab:cross_dataset_fr}
	\resizebox{\linewidth}{!}{%
		\begin{tabular}{@{}lcccccc cccccc@{}}
			\toprule
			\multirow{2}{*}{\textbf{Method}} & \multicolumn{6}{c}{\textbf{LOL-v2-Real}} & \multicolumn{6}{c}{\textbf{LSRW}} \\
			\cmidrule(lr){2-7}\cmidrule(lr){8-13}
			& PSNR$\uparrow$ & SSIM$\uparrow$ & LPIPS$\downarrow$ & DISTS$\downarrow$ & ExpErr$\downarrow$ & ColorErr$\downarrow$
			& PSNR$\uparrow$ & SSIM$\uparrow$ & LPIPS$\downarrow$ & DISTS$\downarrow$ & ExpErr$\downarrow$ & ColorErr$\downarrow$ \\
			\midrule
			ASID & 17.879 & 0.6683 & 0.4921 & 0.2721 & 0.0746 & 0.2264 & 17.293 & 0.5173 & 0.6048 & 0.2607 & 0.0613 & 0.1936 \\
			CATANet & 17.488 & 0.6536 & 0.4567 & 0.2710 & 0.0861 & 0.2641 & 16.922 & 0.5127 & 0.5906 & 0.2650 & 0.0816 & 0.2580 \\
			HiT-SR & 18.943 & \textbf{0.6985} & 0.4213 & 0.2863 & 0.0854 & 0.2667 & 16.174 & 0.5147 & 0.6061 & 0.2879 & 0.0908 & 0.2940 \\
			PFT-SR & 18.927 & 0.6826 & 0.5298 & 0.2966 & 0.0844 & 0.2577 & 17.463 & 0.5098 & 0.6545 & 0.2833 & 0.0767 & 0.2378 \\
			DHGM & 19.103 & 0.6189 & 0.4914 & 0.2789 & 0.0597 & 0.1866 & 17.383 & 0.5192 & 0.5036 & 0.2611 & \textbf{0.0561} & \textbf{0.1805} \\
			\textbf{DeLang-SR} & \textbf{19.512} & 0.6973 & \textbf{0.4060} & \textbf{0.2382} & \textbf{0.0550} & \textbf{0.1749} & \textbf{17.639} & \textbf{0.6579} & \textbf{0.4941} & \textbf{0.2137} & 0.0611 & 0.1868 \\
			\bottomrule
		\end{tabular}%
	}
\end{table}

\begin{table}[!h]
	\centering
	\setlength{\tabcolsep}{4.5pt}
	\renewcommand{\arraystretch}{0.96}
	\caption{No-reference perceptual evaluation under the same cross-dataset protocol.}
	\label{tab:cross_dataset_nr}
	\resizebox{0.84\linewidth}{!}{%
		\begin{tabular}{@{}lccc ccc@{}}
			\toprule
			\multirow{2}{*}{\textbf{Method}} & \multicolumn{3}{c}{\textbf{LOL-v2-Real}} & \multicolumn{3}{c}{\textbf{LSRW}} \\
			\cmidrule(lr){2-4}\cmidrule(lr){5-7}
			& MUSIQ$\uparrow$ & MANIQA$\uparrow$ & CLIP-IQA$\uparrow$ & MUSIQ$\uparrow$ & MANIQA$\uparrow$ & CLIP-IQA$\uparrow$ \\
			\midrule
			ASID & 38.078 & 0.2155 & 0.1925 & 38.064 & 0.2155 & 0.1684 \\
			CATANet & 41.756 & 0.2088 & 0.1801 & 39.027 & 0.2008 & 0.1489 \\
			HiT-SR & 32.061 & 0.2014 & 0.1458 & 33.438 & 0.2336 & 0.1508 \\
			PFT-SR & 28.927 & 0.1770 & 0.1896 & 31.787 & 0.2285 & 0.1741 \\
			DHGM & 43.575 & 0.2554 & 0.1919 & 43.127 & 0.2247 & 0.1667 \\
			\textbf{DeLang-SR} & \textbf{55.826} & \textbf{0.2976} & \textbf{0.3755} & \textbf{54.865} & \textbf{0.2846} & \textbf{0.3434} \\
			\bottomrule
		\end{tabular}%
	}
\end{table}

On LOL-v2-Real, DeLang-SR obtains the best PSNR, LPIPS, DISTS, exposure error, and color error among the evaluated methods, while its SSIM is within $0.0012$ of the best result. It also achieves the highest scores on all three no-reference metrics. On LSRW, DeLang-SR performs best in PSNR, SSIM, LPIPS, DISTS, MUSIQ, MANIQA, and CLIP-IQA, whereas DHGM yields lower exposure and color errors. These results indicate that the fidelity--perception improvements transfer across the two constructed dark-SR protocols, while the illumination-related trade-off remains dataset dependent.

\subsection{Update-Strength Bounds and Spatial Variation}
\label{app:uniform_strength}

We vary the output bounds at inference time using the same 5K checkpoint, without re-optimizing the model. Across the tested intervals in Table~\ref{tab:sigma_sensitivity}, PSNR varies by $0.106$ dB, SSIM by $0.0013$, and LPIPS by $0.0069$. The default interval \([0.15,0.95]\) gives the best PSNR and LPIPS, whereas alternative intervals yield small improvements in exposure or color error.

\begin{table}[!h]
	\centering
	\setlength{\tabcolsep}{5.0pt}
	\renewcommand{\arraystretch}{0.96}
	\caption{Inference-time sensitivity to \((\sigma_{\min},\sigma_{\max})\) on RELLISUR \(\times2\) at 5K steps.}
	\label{tab:sigma_sensitivity}
	\resizebox{0.82\linewidth}{!}{%
		\begin{tabular}{@{}lccccc@{}}
			\toprule
			Bounds & PSNR$\uparrow$ & SSIM$\uparrow$ & LPIPS$\downarrow$ & ExpErr$\downarrow$ & ColorErr$\downarrow$ \\
			\midrule
			$[0.10,0.90]$ & 22.910 & 0.7588 & 0.2536 & 0.0374 & 0.1148 \\
			$[0.15,0.85]$ & 22.880 & 0.7596 & 0.2551 & 0.0341 & 0.1065 \\
			$[0.15,0.95]$ (default) & \textbf{22.986} & 0.7594 & \textbf{0.2482} & 0.0360 & 0.1120 \\
			$[0.20,0.90]$ & 22.930 & \textbf{0.7601} & 0.2518 & 0.0349 & 0.1082 \\
			\bottomrule
		\end{tabular}%
	}
\end{table}

To isolate spatial variation, we use the same 5K checkpoint at inference time, without retraining, and replace each predicted field with its image-wise mean before broadcasting the resulting scalar to all latent locations. This control preserves the average update magnitude for each image while removing its spatial allocation. As shown in Table~\ref{tab:uniform_strength}, the spatial field improves PSNR by $0.243$ dB and SSIM by $0.0053$, and reduces LPIPS by $0.0125$. ExpErr and ColorErr change by only $0.0002$ and $0.0015$, respectively, indicating that the main benefit arises from spatially allocating the update rather than changing its global mean.

\begin{table}[!h]
	\centering
	\setlength{\tabcolsep}{5.2pt}
	\renewcommand{\arraystretch}{0.96}
	\caption{Ablation of spatial variation in the update-strength field on RELLISUR \(\times2\) at 5K steps.}
	\label{tab:uniform_strength}
	\resizebox{0.78\linewidth}{!}{%
		\begin{tabular}{@{}lccccc@{}}
			\toprule
			Update strength & PSNR$\uparrow$ & SSIM$\uparrow$ & LPIPS$\downarrow$ & ExpErr$\downarrow$ & ColorErr$\downarrow$ \\
			\midrule
			Uniform image-wise mean & 22.743 & 0.7541 & 0.2607 & \textbf{0.0358} & \textbf{0.1105} \\
			\textbf{Spatial field} & \textbf{22.986} & \textbf{0.7594} & \textbf{0.2482} & 0.0360 & 0.1120 \\
			\bottomrule
		\end{tabular}%
	}
\end{table}

\subsection{Additional Update-Strength Field Visualizations}
\label{app:update_strength_visual}

\begin{figure}[!t]
	\centering
	\includegraphics[width=0.98\linewidth]{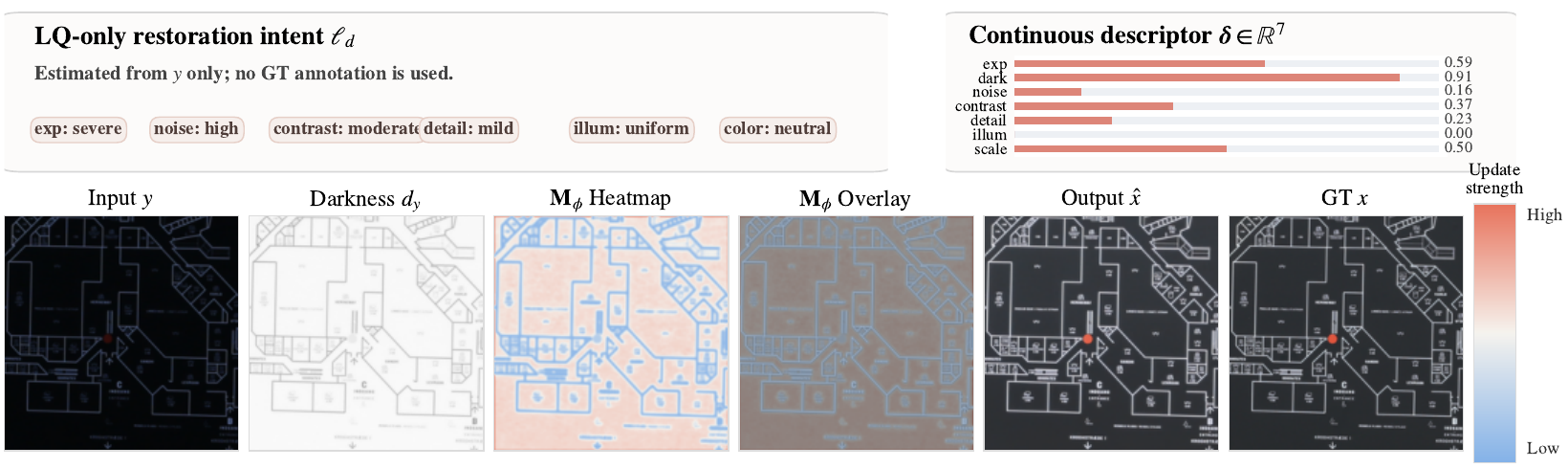}
	\vspace{-0.15cm}
	
	\includegraphics[width=0.98\linewidth]{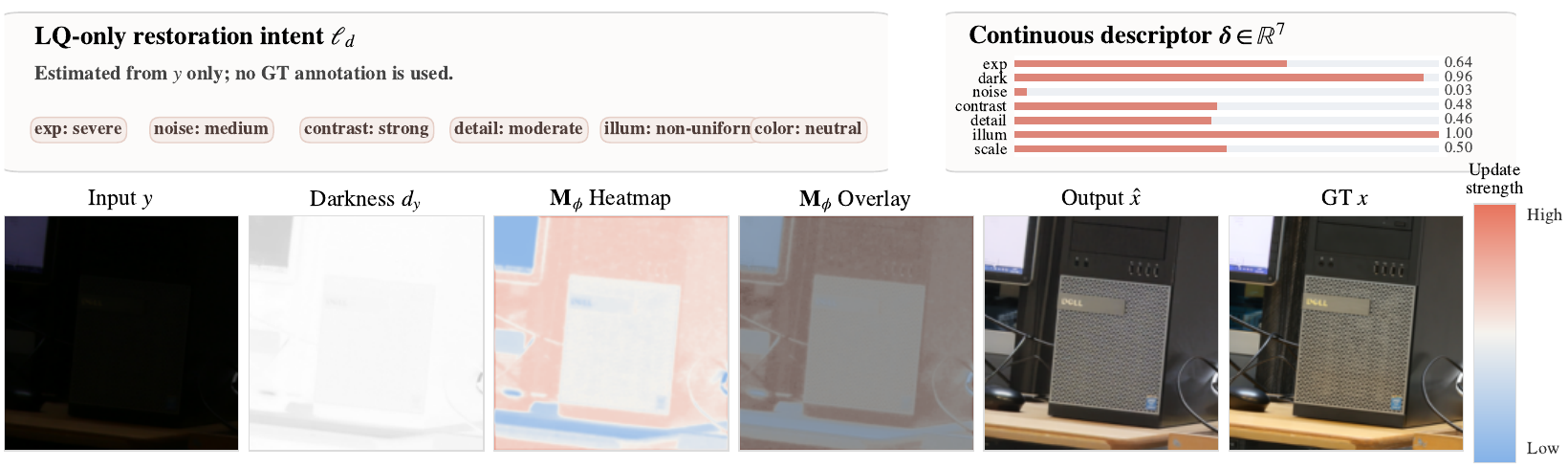}
	\vspace{-0.15cm}
	
	\includegraphics[width=0.98\linewidth]{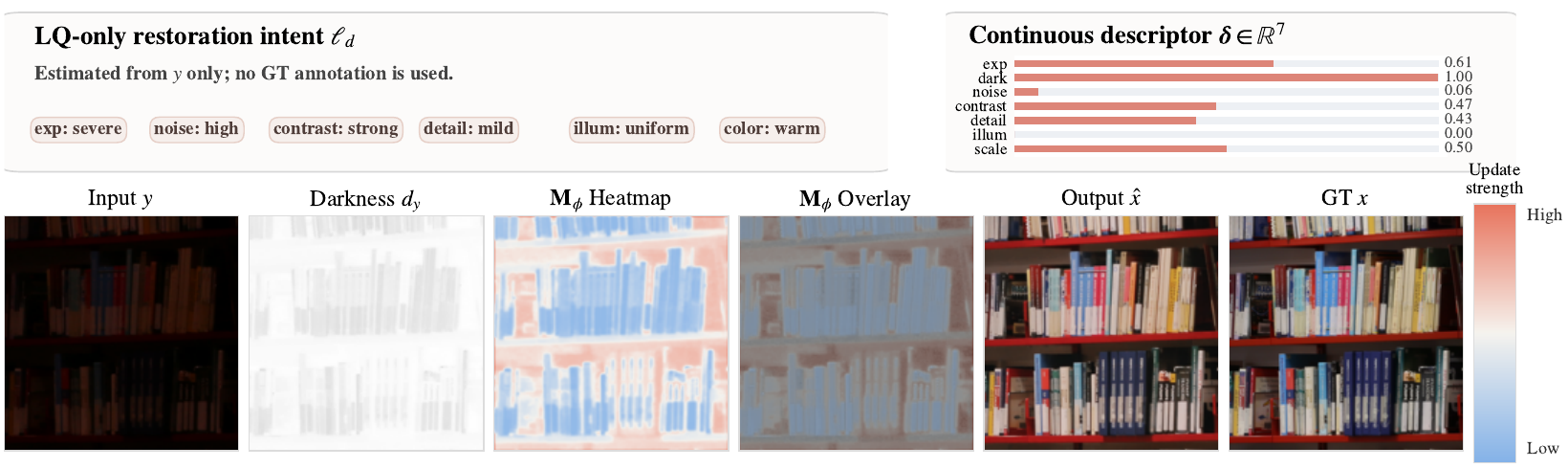}
	\vspace{-0.15cm}
	
	\includegraphics[width=0.98\linewidth]{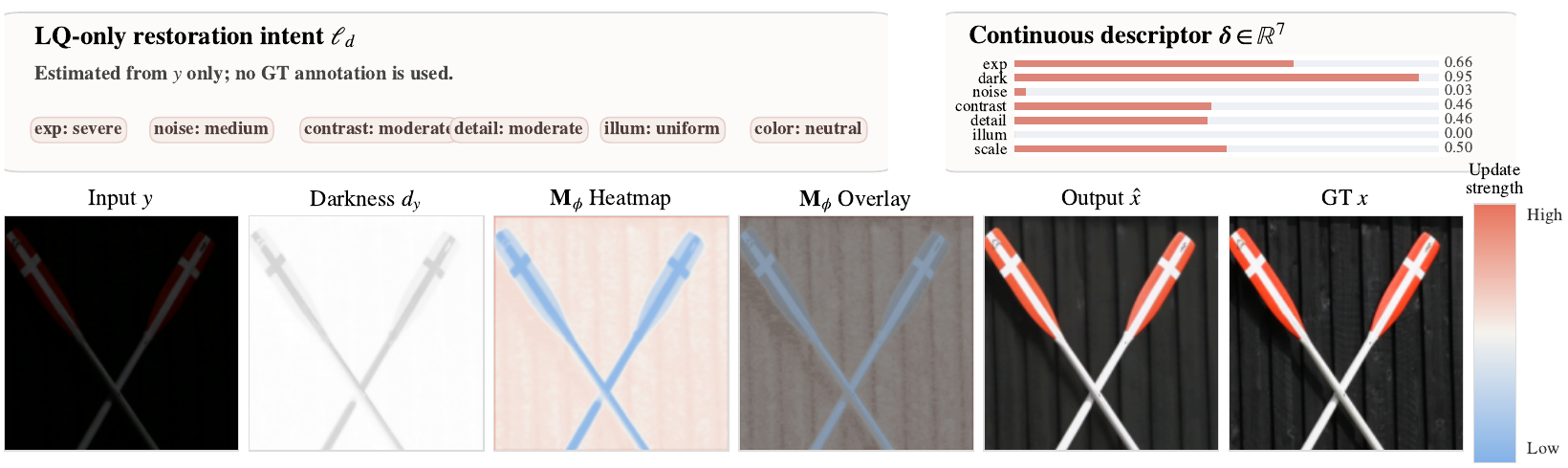}
	\vspace{-0.2cm}
	\caption{
		Additional visualizations of the update-strength field \(\mathbf{M}_{\phi}(y,\boldsymbol{\delta})\).
		For each example, \(\ell_d\) and \(\boldsymbol{\delta}\) are estimated from the input \(y\) without GT annotations.
		The predicted maps provide spatially non-uniform step-size fields for one-step SD3 flow restoration.
	}
	\label{fig:app_update_strength_maps}
	\vspace{-0.2cm}
\end{figure}

We provide additional visualizations of the degradation-aware update-strength field 
\(\mathbf{M}_{\phi}(y,\boldsymbol{\delta})\) in Fig.~\ref{fig:app_update_strength_maps}. 
For each case, the LQ-only restoration intent \(\ell_d\) and continuous descriptor \(\boldsymbol{\delta}\) are estimated from the input \(y\) without using the paired GT. 
The predicted update-strength fields are spatially non-uniform and provide interpretable local step-size maps for the one-step SD3 flow restoration. 
These examples further illustrate that DeLang-SR does not apply a uniform global update, but adapts the local update magnitude according to degradation severity and visual evidence.

\begin{figure}[!t]
	\centering
	\includegraphics[width=0.98\linewidth]{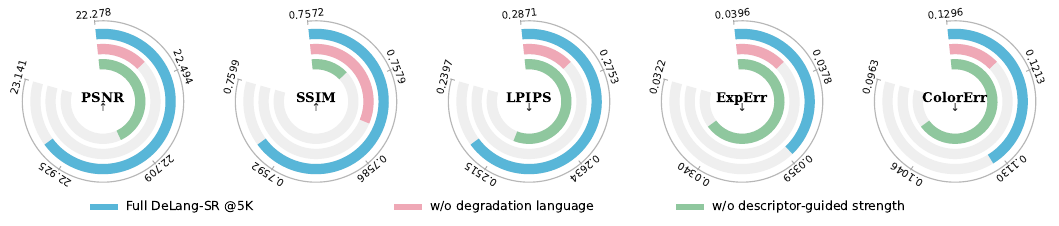}
	\vspace{-0.2cm}
	\caption{
		Additional ring-chart visualization of selected component ablations on \textbf{\textit{RELLISUR}} \(\times2\).
		The chart compares Full DeLang-SR, w/o degradation language, and w/o descriptor conditioning in \(\mathbf{M}_{\phi}\) across PSNR, SSIM, LPIPS, ExpErr, and ColorErr.
		The full model achieves a better overall trade-off, supporting the complementarity of direction steering and magnitude control.
	}
	\label{fig:app_component_ablation_ring}
	\vspace{-0.2cm}
\end{figure}

\subsection{Inference-Time Condition Source Analysis}
\label{app:condition_source}

\begin{table}[!h]
	\centering
	\setlength{\tabcolsep}{4.2pt}
	\renewcommand{\arraystretch}{1.02}
	\vspace{-0.1cm}
	\caption{
		Inference-time condition source analysis on \textbf{\textit{RELLISUR}} \(\times2\).
		Comparable results across generic, LQ-only, and oracle intents show that DeLang-SR does not rely on oracle degradation annotations.
	}
	\label{tab:condition_source}
	\vspace{-0.25cm}
	\resizebox{0.90\linewidth}{!}{%
		\begin{tabular}{@{}lcccccc@{}}
			\toprule[1.1pt]
			\rowcolor{condhead}
			\textbf{\emph{Condition Source}}
			& \textbf{Uses GT?}
			& \textbf{PSNR\(\uparrow\)}
			& \textbf{SSIM\(\uparrow\)}
			& \textbf{LPIPS\(\downarrow\)}
			& \textbf{ExpErr\(\downarrow\)}
			& \textbf{ColorErr\(\downarrow\)} \\
			\midrule
			
			\rowcolor{condgeneric}
			Generic prompt
			& \nogt & 22.984 & 0.7593 & 0.2487 & 0.0361 & 0.1120 \\
			
			\rowcolor{conddeploy}
			\textbf{LQ-only degradation intent}
			& \nogt & 22.985 & 0.7594 & 0.2482 & 0.0361 & 0.1120 \\
			
			\rowcolor{condoracle}
			Oracle degradation intent
			& \yesgt & 22.985 & 0.7593 & 0.2479 & 0.0360 & 0.1119 \\
			
			\bottomrule[1.1pt]
		\end{tabular}
	}
	\vspace{-0.2cm}
\end{table}

\begin{figure}[!h]
	\centering
	\includegraphics[width=\linewidth,trim=1 2 1 5, clip]{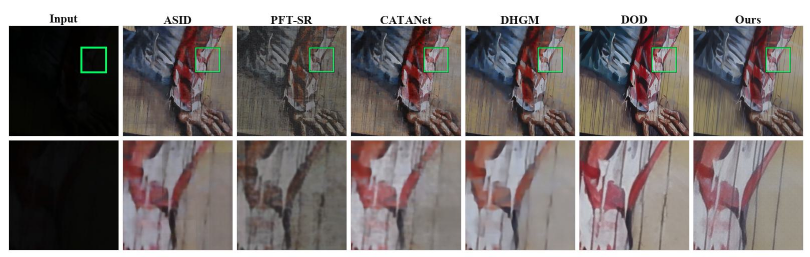}
	\includegraphics[width=\linewidth,trim=1 2 1 5, clip]{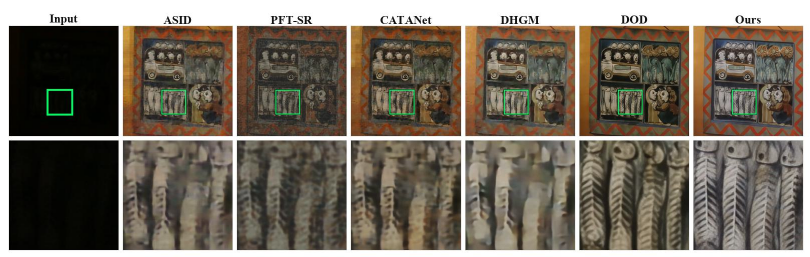}
	\includegraphics[width=\linewidth,trim=1 2 1 5, clip]{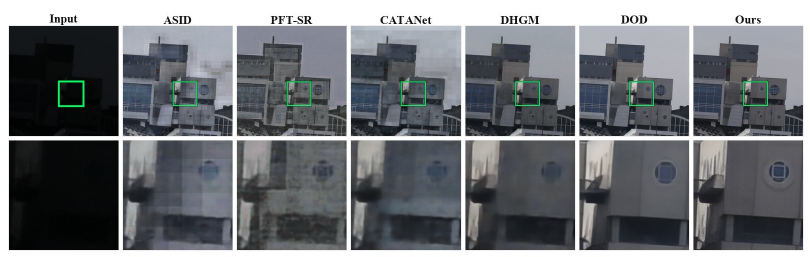}
	\includegraphics[width=\linewidth,trim=1 2 1 5, clip]{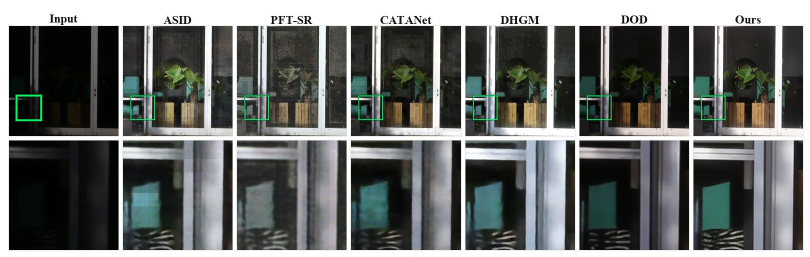}
	\includegraphics[width=\linewidth,trim=1 2 1 5, clip]{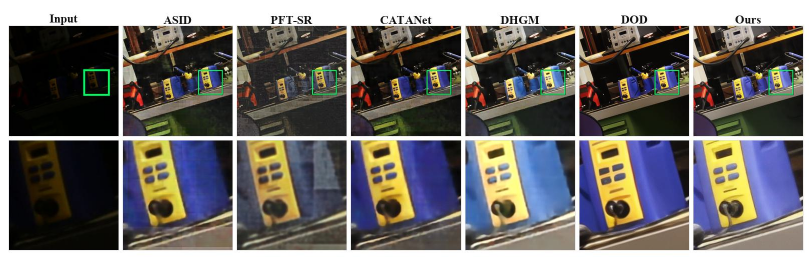}
	\vspace{-0.4cm}
	\caption{
		Additional visual comparison on \textbf{\textit{RELLISUR}} \(\times2\).
		Each row shows the restored full image with the selected ROI marked in green and the corresponding enlarged crop.
		DeLang-SR preserves clearer local structures and more faithful contrast under challenging dark \(\times2\) super-resolution.
	}
	\label{fig:app_rellisur_x2_visual}
	\vspace{-0.2cm}
\end{figure}

\begin{figure}[!h]
	\centering
	\includegraphics[width=\linewidth,trim=1 2 1 5, clip]{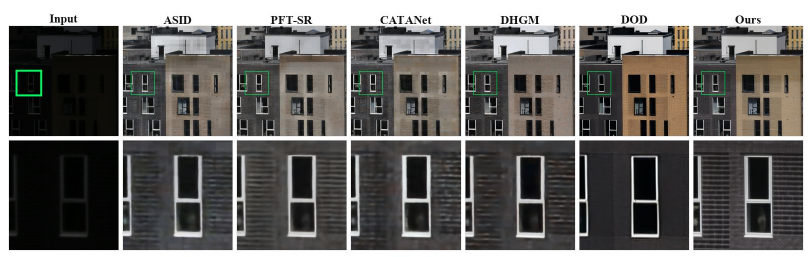}
	\includegraphics[width=\linewidth,trim=1 2 1 5, clip]{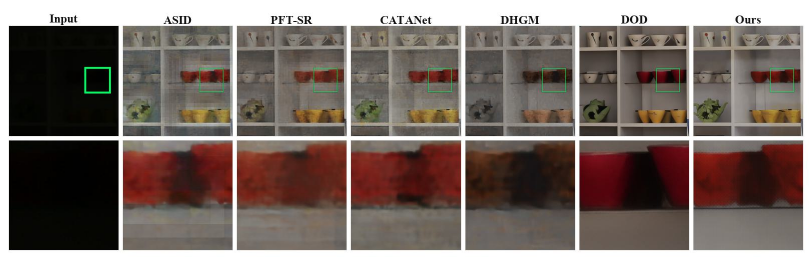}
	\includegraphics[width=\linewidth,trim=1 2 1 5, clip]{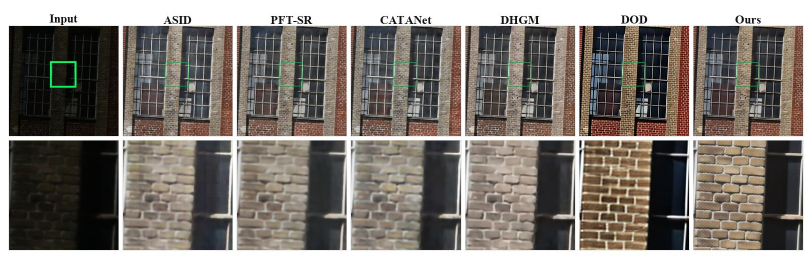}
	\includegraphics[width=\linewidth,trim=1 2 1 5, clip]{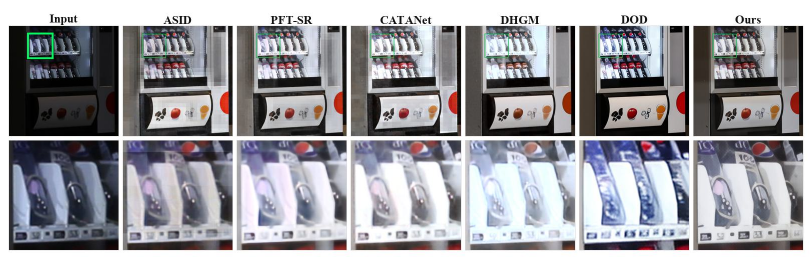}
	\includegraphics[width=\linewidth,trim=1 2 1 5, clip]{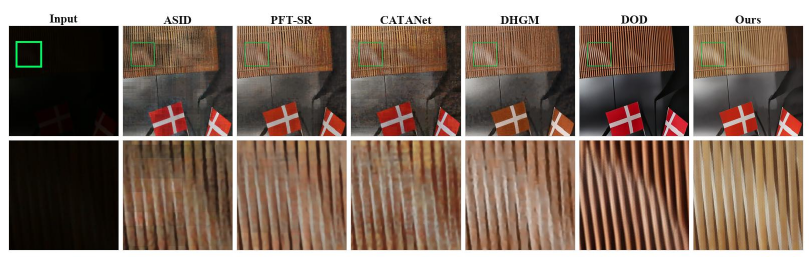}
	\vspace{-0.4cm}
	\caption{
		Additional visual comparison on \textbf{\textit{RELLISUR}} \(\times4\).
		Each row shows the full restored image and the enlarged ROI crop.
		DeLang-SR preserves clearer local structures and more faithful contrast under challenging dark \(\times4\) super-resolution.
	}
	\label{fig:app_rellisur_x4_visual}
	\vspace{-0.2cm}
\end{figure}

We further report the numerical results of the inference-time condition source analysis in Table~\ref{tab:condition_source}. 
The same full checkpoint is evaluated with three condition sources: a fixed generic prompt, the proposed LQ-only degradation intent, and an oracle degradation intent constructed from paired data. 
The comparable performance across these settings shows that DeLang-SR does not rely on oracle degradation annotations during deployment, and that the generated degradation language should be interpreted as a coarse restoration intent rather than a precise label. The generic prompt differs from the LQ-only intent by only $0.001$ dB PSNR and $0.0005$ LPIPS, indicating that detailed sentence composition has a limited effect on these aggregate metrics. This comparison changes only the language condition; all variants retain the continuous descriptor \(\boldsymbol{\delta}\) and the visual-token pathway, so it does not remove the complete input-statistics block.

\subsection{Component Ablation Visualization}
\label{app:component_ablation_visual}

We provide an additional ring-chart visualization of selected component ablations in Fig.~\ref{fig:app_component_ablation_ring}. 
This figure complements the quantitative ablation table in the main paper by comparing the full DeLang-SR model with two representative variants: removing degradation language and removing descriptor conditioning from the update-strength predictor. 
Across fidelity, perceptual, exposure, and color-related metrics, the full model generally provides a better overall trade-off. 
This supports the complementarity between degradation-language direction steering and descriptor-guided magnitude control.

\subsection{Additional Visual Results on RELLISUR}
\label{app:rellisur_visual}

We provide additional qualitative comparisons on RELLISUR under both \(\times2\) and \(\times4\) settings in Figs.~\ref{fig:app_rellisur_x2_visual} and~\ref{fig:app_rellisur_x4_visual}. 
For each example, we show the restored full image with the selected ROI marked in green and the corresponding enlarged crop. 
The \(\times2\) results examine local structure and contrast restoration under dark degradation, while the \(\times4\) results further test fine-detail recovery from more limited spatial observations. 
Across both scaling factors, DeLang-SR preserves clearer local structures, repetitive textures, and natural contrast, whereas competing methods often suffer from over-smoothing, color shifts, or unstable texture reconstruction under severe darkness.

\end{document}